\documentclass[sigconf]{acmart}
\usepackage{bm}
\usepackage{amsfonts}
\usepackage{multirow}
\usepackage{booktabs}
\usepackage{mathrsfs}
\usepackage[ruled]{algorithm2e}

\DeclareMathOperator*{\ie}{i.e.,}
\DeclareMathOperator*{\eg}{e.g.,}
\setcitestyle{numbers,sort&compress}

\AtBeginDocument{%
  \providecommand\BibTeX{{%
    \normalfont B\kern-0.5em{\scshape i\kern-0.25em b}\kern-0.8em\TeX}}}

\copyrightyear{2021}
\acmYear{2021}
\setcopyright{acmlicensed}\acmConference[MM '21]{Proceedings of the 29th
	ACM International Conference on Multimedia}{October 20--24, 2021}{Virtual
	Event, China}
\acmBooktitle{Proceedings of the 29th ACM International Conference on
	Multimedia (MM '21), October 20--24, 2021, Virtual Event, China}
\acmPrice{15.00}
\acmISBN{978-1-4503-8651-7/21/10}
\acmDOI{10.1145/3474085.xxxxxxx}

\begin{document}
\fancyhead{}
\title{When Video Classification Meets Incremental Classes}

\author{Hanbin Zhao}
\affiliation{
	\institution{College of Computer Science and Technology}
	\country{Zhejiang University}
	\city{Hangzhou}}
\email{zhaohanbin@zju.edu.cn}

\author{Xin Qin}
\affiliation{
	\institution{College of Computer Science and Technology}
	\country{Zhejiang University}
	\city{Hangzhou}}
\email{xinqin@zju.edu.cn}

\author{Shihao Su}
\affiliation{
	\institution{College of Computer Science and Technology}
	\country{Zhejiang University}
	\city{Hangzhou}}
\email{shihaocs@zju.edu.cn}

\author{Yongjian Fu}
\affiliation{
	\institution{College of Computer Science and Technology}
	\country{Zhejiang University}
	\city{Hangzhou}}
\email{yjfu@zju.edu.cn}

\author{Zibo Lin}
\affiliation{
	\institution{College of Computer Science and Technology}
	\country{Zhejiang University}
	\city{Hangzhou}}
\email{zibolin@zju.edu.cn}

\author{Xi Li}
\authornote{Corresponding author}
\affiliation{
	\institution{College of Computer Science and Technology}
	\country{Zhejiang University}
	\city{Hangzhou}}
\email{xilizju@zju.edu.cn}

\renewcommand{\shortauthors}{Zhao, et al.}

\begin{abstract}
	With the rapid development of social media, tremendous videos with new classes are generated daily, which raise an urgent demand for video classification methods that can continuously update new classes while maintaining the knowledge of old videos with limited storage and computing resources. In this paper, we summarize this task as \textit{Class-Incremental Video Classification (CIVC)} and propose a novel framework to address it. 
	As a subarea of incremental learning tasks, the challenge of \textit{catastrophic forgetting} is unavoidable in CIVC.
    To better alleviate it, we utilize some characteristics of videos. First, we decompose the spatio-temporal knowledge before distillation rather than treating it as a whole in the knowledge transfer process; trajectory is also used to refine the decomposition.
	Second, we propose a dual granularity exemplar selection method to select and store representative video instances of old classes and key-frames inside videos under a tight storage budget. 
	We benchmark our method and previous SOTA class-incremental learning methods on Something-Something V2 and Kinetics datasets, and our method outperforms previous methods significantly.
\end{abstract}

%



\maketitle

\section{Introduction}
\begin{figure}[t]
	\begin{center}
		\includegraphics[width=0.7\linewidth]{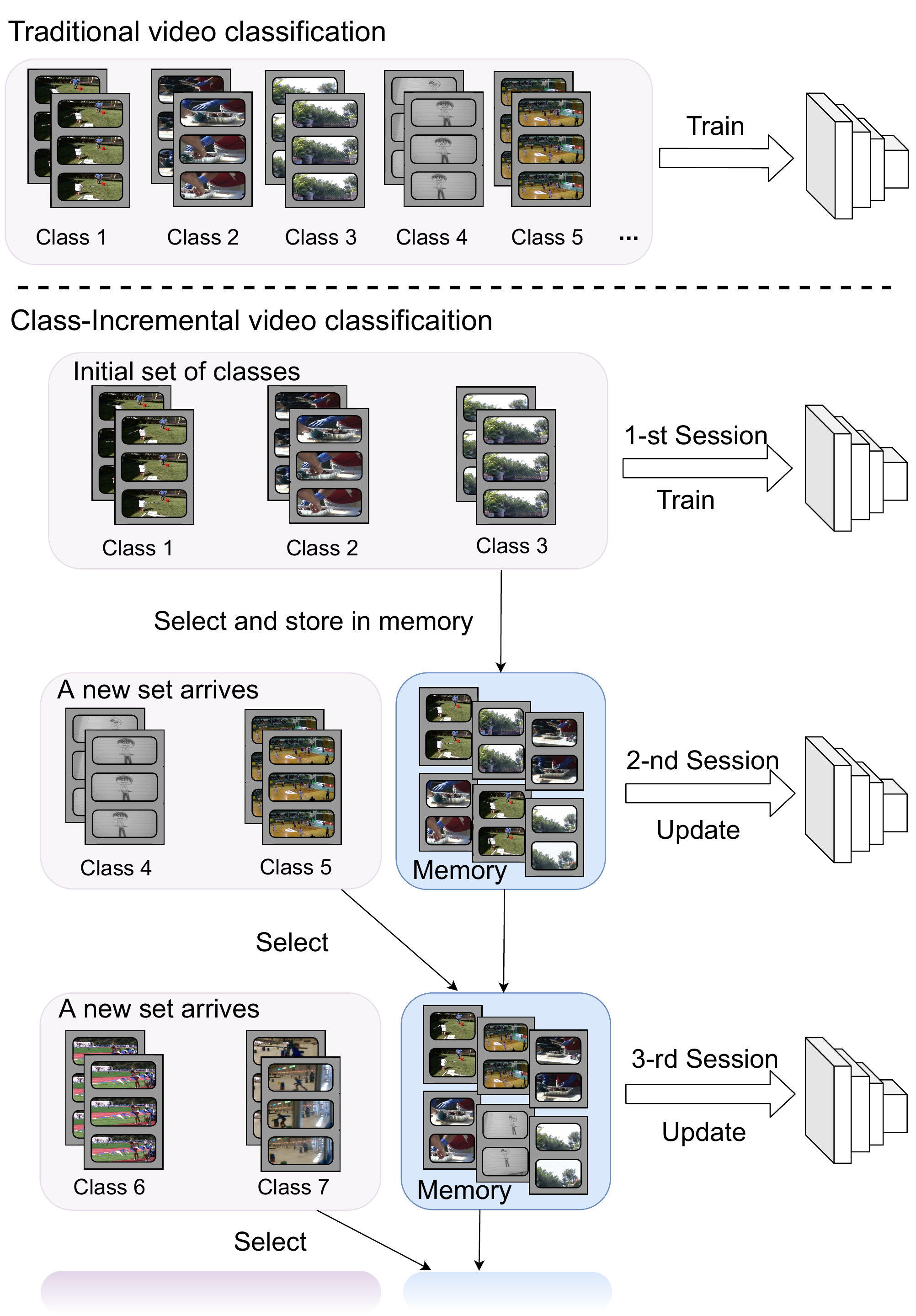}
	\end{center}
	\caption{Comparison of traditional Video Classification and Class-Incremental Video Classification (CIVC). In traditional Video Classification, all videos are available during training. However, in CIVC, new videos of new classes arrive sequentially. The model has to be updated with new videos while maintaining performance for old data.}
	\label{fig:one}
	\vspace{-9pt}
\end{figure}

\begin{figure*}[t]
	\begin{center}
		\includegraphics[width=0.23\linewidth]{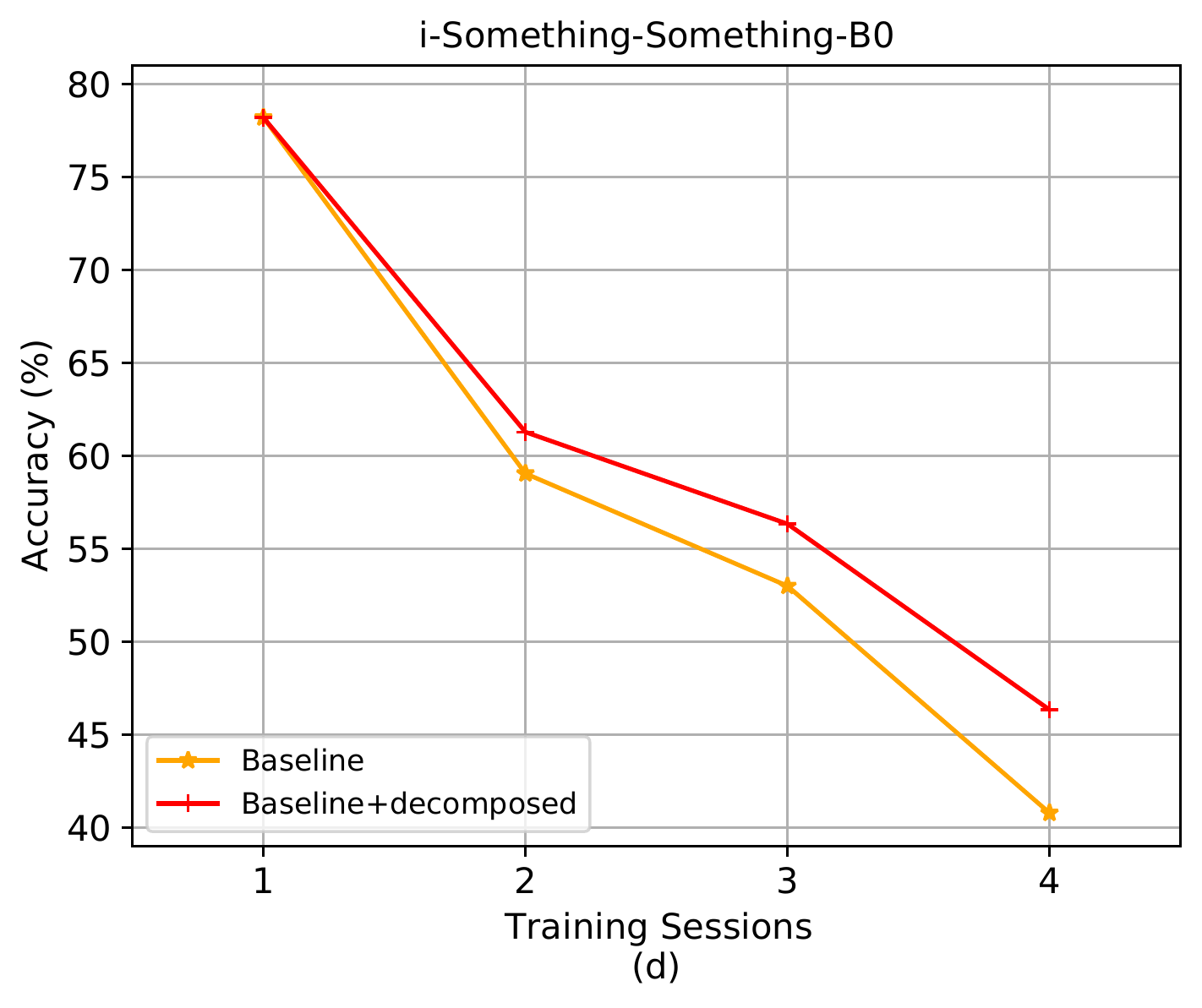}
		\includegraphics[width=0.23\linewidth]{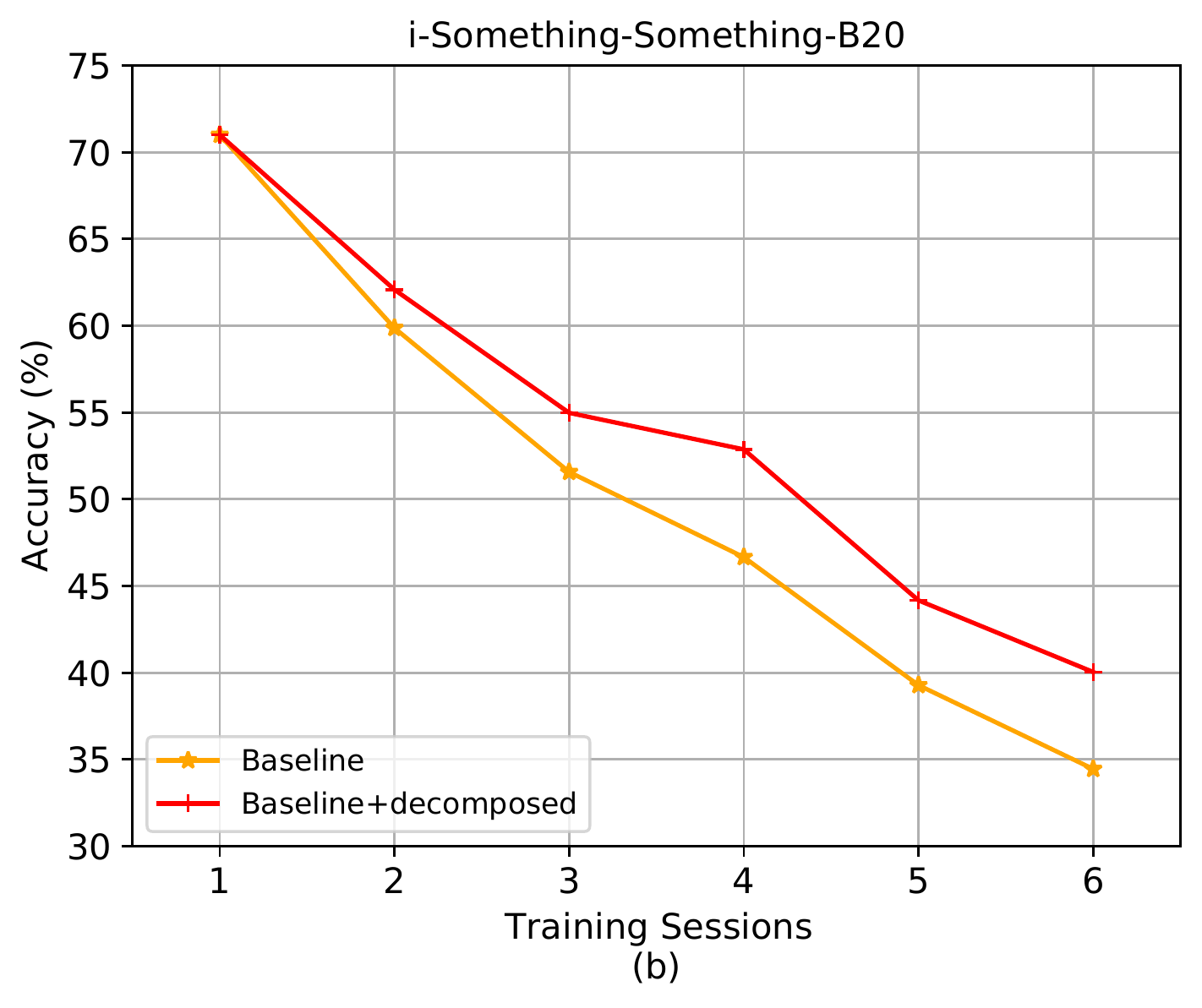}
		\includegraphics[width=0.23\linewidth]{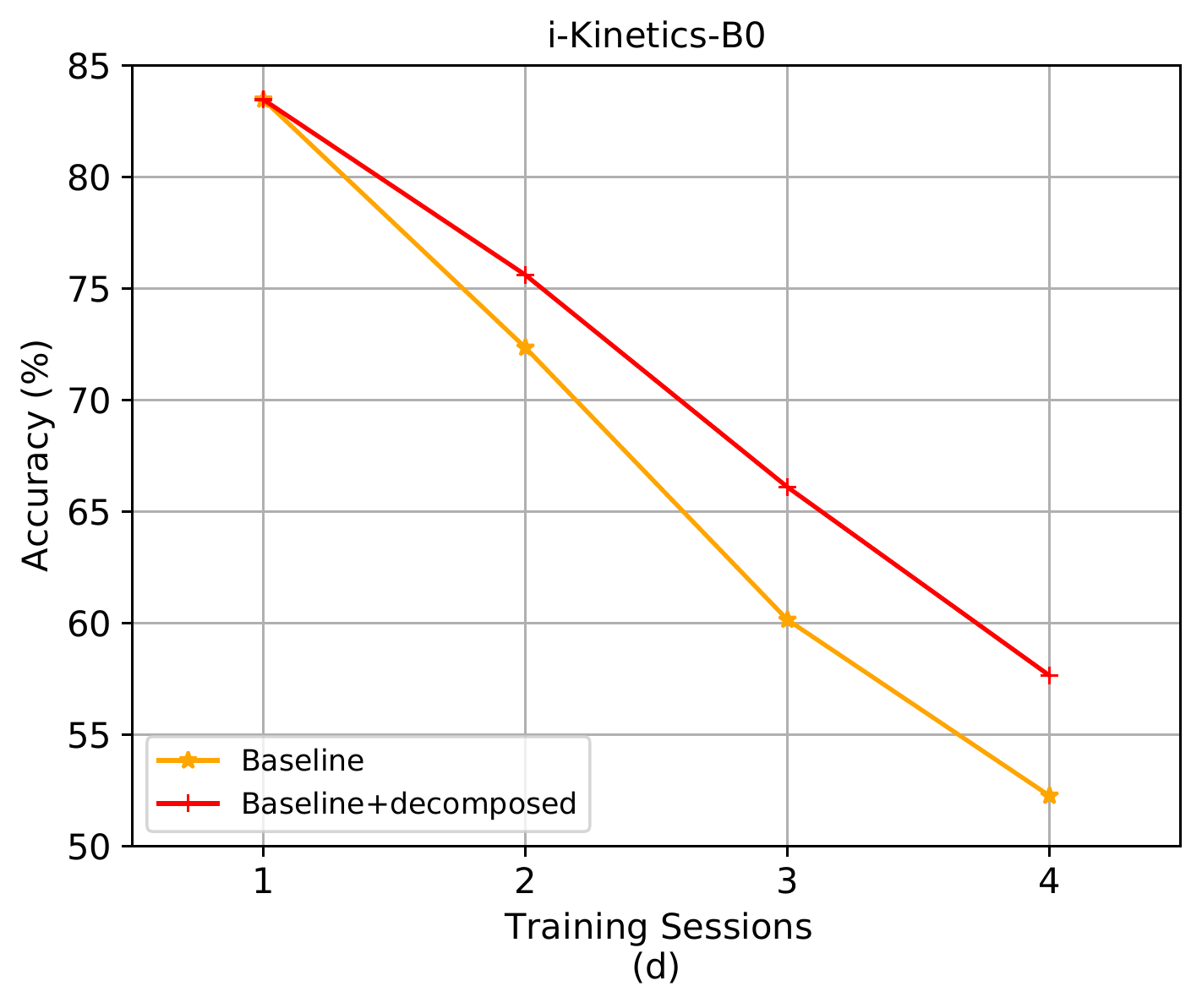}
		\includegraphics[width=0.23\linewidth]{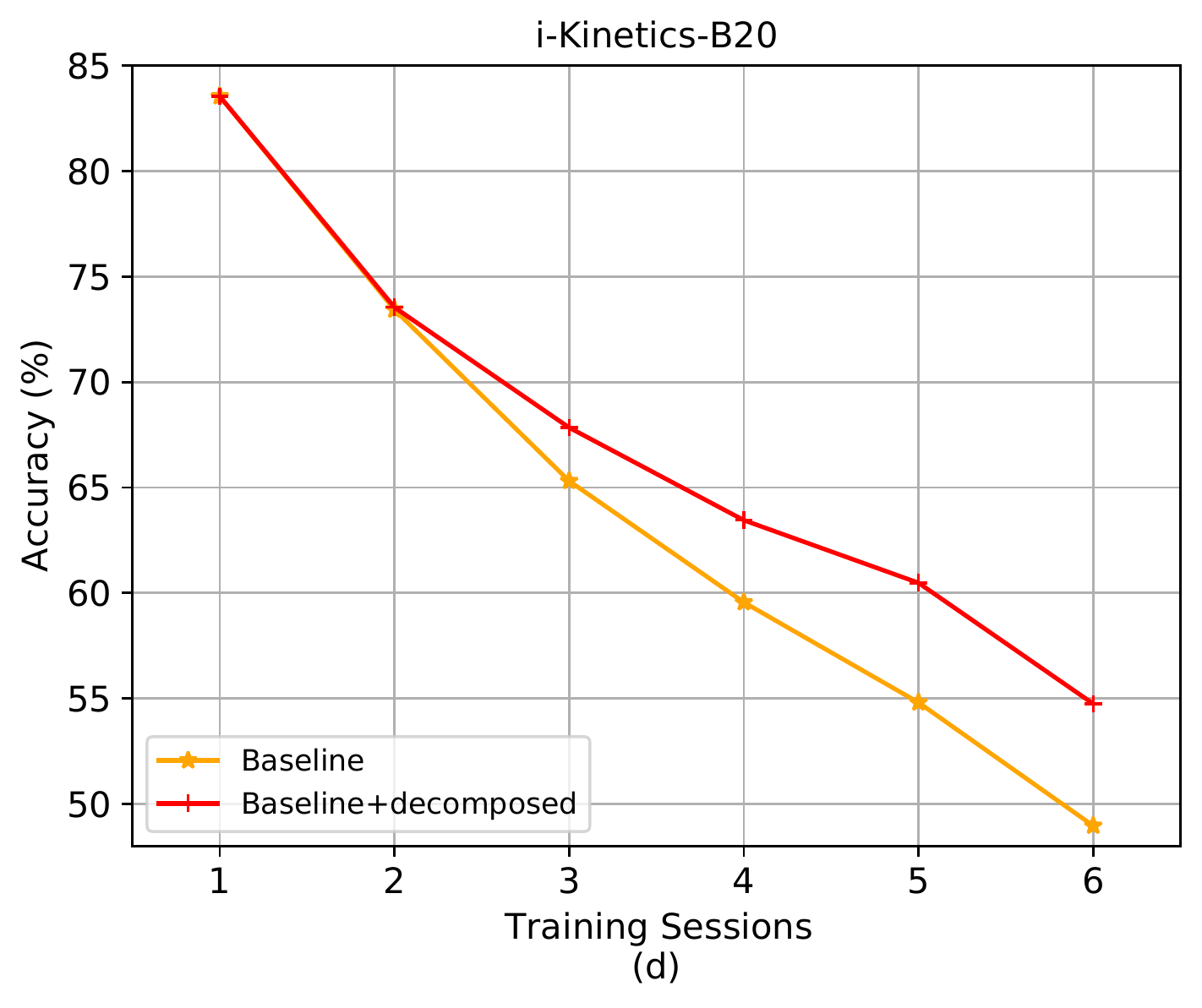}
	\end{center}
	\caption{Comparison of directly distilling the spatio-temporal feature ($\ie$ Baseline) and distilling the spatial and temporal feature separately ($\ie$ Baseline+decomposed) on four CIVC benchmarks. A set of new-class videos arrives at each training session and the model is updated on it. We calculate the accuracy on the test part data of seen classes. Baseline+decomposed achieves better performance than Baseline at each session.}
	\label{fig:two}
	\vspace{-8pt}
\end{figure*}

Recent years have witnessed an immense explosion of multimedia data, especially videos, due to the rapid development of social network applications ($\eg$ \textit{Instagram}, \textit{TikTok}); meanwhile, deep learning boosts the development of video understanding~\cite{lin2019tsm, cao2020few, tran2019video, xie2018rethinking}. 
Previous works on video understanding and analysis achieve outstanding performance with the fixed number of video classes during training. 
However, it is usually not the case in real-world applications where new video classes emerge continuously. 
For example, there are millions of short videos, along with new classes ($\eg$ \textit{themes or tags}), shared by users in social media daily. 
The video understanding model is required to adapt to those new classes.

An intuitive solution to recognize all classes is to mix all videos from both old and new classes and train a new model from scratch. 
Unfortunately, it is infeasible to train a model from scratch repeatedly every time the new classes of videos emerge as the computation overhead is unaffordable. 
Also, maintaining such a huge dataset for training raises a great challenge in storage resources.
An alternative is to fine-tune the old model using videos of new classes. 
In this way, the model achieves good performance on new video classes; however, a significant performance drop on old classes occurs (such a phenomenon is called \textit{catastrophic forgetting}~\cite{mccloskey1989catastrophic}). 
Therefore, there is a significant interest of research in training a model with high performance on both old and new videos on a tight budget in both storage and computation resources. 
Among those video understanding tasks, we focus on video classification in this paper, and thus we name this problem as: \textit{Class-Incremental Video Classification} (CIVC) (as shown in Figure~\ref{fig:one}).

CIVC, as a first proposed task, is related to Class-Incremental Image Classification (CIIC), which is a widely explored area. 
In CIIC, previous methods tackle the forgetting problem in two ways: 1) transferring the old-class knowledge while learning the new-class knowledge, usually involved with distillation, and 2) utilizing a limited memory to select and store a few most representative old-class images. 
In contrast to images that contain only spatial information, videos are much more complicated as they convey intricate spatio-temporal information; meanwhile, videos comprise a sequence of frames, each of which is an image. 
Hence, based on the above two characteristics of videos, we improve the CIIC method to better fit CIVC task according to the following two strategies: 1) the decomposed spatial-temporal knowledge transfer mechanism, where the spatial-temporal knowledge is decomposed and then distilled separately; 2) the dual granularity exemplars selection, which not only selects exemplars at video granularity but also selects the key-frames inside video exemplars, $\ie$ inner-video granularity.

First, we develop an effective knowledge transfer strategy handling the spatio-temporal knowledge conveyed in videos.
Due to the different characteristics of spatial information and temporal information, we observe that directly distilling the fused spatio-temporal feature transfers less knowledge than distilling the spatial and temporal feature separately (as shown in Figure~\ref{fig:two}). 
Hence, we claim that the spatial and temporal feature should be decomposed before distillation.
A naive decomposition by pooling on different spatio-temporal dimensions has achieved improvement on performance by a tolerable margin; utilizing trajectory, which is considered as key temporal knowledge in videos~\cite{wang2015action, zhao2018trajectory}, to refine the decomposition achieves a better distillation quality.

Second, we propose to select the most representative samples at two granularity ($\ie$ instance-granularity and inner-instance-granularity).
As mentioned above, a video is a sequence of images, where some images that play remarkable roles in the temporal sequence can be selected as key-frames. 
Therefore, besides sampling at video instance-granularity, we also select key-frames from representative videos base on the temporal information. 
By preserving the most informative frames in the most representative videos, the memory budget for storing the videos is further compressed.

In order to acquire a more convincing assessment, we build the CIVC baselines by adapting the state-of-the-art CIIC methods to this new task and compare our method with them. 
We illustrate the effectiveness of our proposed framework by extensive evaluation on popular video classification benchmarks, Something-Something V2 and Kinetics.

The contributions of this paper can be summarized as follows:
\begin{itemize}
	\item We establish the CIVC task and propose the corresponding framework. We benchmark our method and previous SOTA class-incremental learning methods on two popular video classification datasets under different experimental settings.
	\item We design a novel knowledge transfer approach that ingeniously decomposed the fused spatio-temporal feature with the help of trajectory before distillation. The decomposition has a significant performance improvement.
	\item We design a simple yet effective dual granularity exemplars selecting strategy that selects and stores the key-frames of most representative video exemplars. This strategy reduces the memory budget by a large margin.
\end{itemize}


\section{Related Work}
\subsection{Video Classification}
With the rise of deep learning, CNNs dominated the video understanding literature recently~\cite{lin2019tsm, cao2020few, tran2019video, wu2018context, wu2020adaptive, qin2021pcmnet, li2018multi, ji2019human}, instead of hand-crafted features~\cite{DBLP:conf/cvpr/WangKSL11,wang2013action,DBLP:conf/eccv/PengZQP14,DBLP:conf/cvpr/LanLLHR15,10.1145/2393347.2396511}. Since video understanding intuitively needs motion information, finding an appropriate way to describe the temporal relationship between frames is essential to improving the performance of video classification. Hence, two-stream networks~\cite{DBLP:conf/nips/SimonyanZ14} which include a spatial stream and a temporal stream are proposed. The spatial stream takes raw video frames as input to capture visual appearance information. The temporal stream takes a stack of optical flow images as input to capture motion information between video frames. It is the first time, a CNN-based approach achieved performance similar to the previous best hand-crafted feature IDT~\cite{wang2013action}. After that, a large number of improvements have been made based on this work, $eg$, using deeper networks~\cite{DBLP:conf/cvpr/HaraKS18}, preventing networks from overfitting~\cite{wang2016temporal}, fusing two-stream features better~\cite{DBLP:conf/cvpr/FeichtenhoferPZ16, DBLP:conf/nips/FeichtenhoferPW16, DBLP:conf/cvpr/FeichtenhoferPW17a,DBLP:conf/cvpr/WangLWY17}, and multi-stream network~\cite{DBLP:conf/mm/WuJWYX16}. 

However, pre-computing optical flow is computationally intensive and storage demanding, which is not friendly for large-scale training or real-time deployment. Hence, this leads to the usage of 3D CNN to model the temporal and spatial information together~\cite{DBLP:journals/pami/JiXYY13, DBLP:conf/iccv/TranBFTP15, carreira2017quo}. Although the performance of C3D~\cite{DBLP:conf/iccv/TranBFTP15} on standard benchmarks is not satisfactory, but it shows strong generalization capability. 
I3D~\cite{carreira2017quo} inflates the ImageNet pre-trained 2D model weights to their counterparts in the 3D model, and pre-trained on a new large-scale dataset, Kinetics400, getting the best classification accuracy and pushing this task to the next level. In the next few years, 3D CNNs advanced quickly and became top performers on almost every benchmark dataset~\cite{ DBLP:conf/iccv/Feichtenhofer0M19, DBLP:conf/cvpr/0004GGH18, DBLP:conf/cvpr/Feichtenhofer20}. To reduce the complexity of 3D network training, many works factorize the 3D kernel to two separate operations ($\ie$ a 2D spatial convolution and a 1D temporal convolution)~\cite{DBLP:conf/iccv/QiuYM17, DBLP:conf/cvpr/TranWTRLP18, DBLP:conf/eccv/XieSHTM18}, or mix 2D and 3D convolutions in a single network~\cite{DBLP:conf/eccv/XieSHTM18,DBLP:conf/cvpr/ZhouSZZ18,DBLP:conf/cvpr/WangL0G18,DBLP:conf/eccv/ZolfaghariSB18}. Recently, many works perform temporal modeling with an extra module instead of the simple 1D temporal convolution~\cite{lin2019tsm, DBLP:conf/aaai/ShaoQL20, DBLP:conf/iccv/JiangWGWY19, DBLP:conf/cvpr/LiJSZKW20, DBLP:conf/aaai/LiuLWWTWLHL20}. All the above works focus on the traditional video classification scenario where all classes of data are available when training the models. Our work focuses on the CIVC scenario where new video classes emerges continuously.

\subsection{Class-Incremental Image Classification}
Recently, there has been a large body of research in class-incremental image classification~\cite{delange2021continual, zhao2020mgsvf}. These works mainly tackle the forgetting problem by effective old-class knowledge transfer and utilize a limited memory to keep exemplars of old classes. 
As for effective knowledge transfer, LWF~\cite{li2017learning} first applies knowledge distillation~\cite{hinton2015distilling} to alleviate the forgetting problem. iCaRL~\cite{rebuffi2017icarl} combines the idea and compute the distillation loss on the prediction of the network. EEIL~\cite{castro2018end} extends iCaRL by learning the network and classifier with an end-to-end approach. BiC~\cite{wu2019large} tried to balance the classifier by a bias correction layer, where the layer is trained on a separate validation set. LUCIR~\cite{hou2019learning} also introduces multiple techniques to correct the classifier. PODNet~\cite{douillard2020podnet} utilizes a spatial-based distillation loss to restrict the change of model. TPCIL~\cite{tao2020topology} makes the model preserve the topology of CNN's feature space.
As for keeping the exemplars, iCaRL proposes to replay old knowledge with a handful of herding exemplars. Herding~\cite{welling2009herding} picks the nearest neighbors of the average sample per class. The same herding exemplars are also utilized in multiple works~\cite{iscen2020memory,castro2018end,wu2019large,hou2019learning,zhao2021memory}.
Some methods~\cite{shin2017continual, kamra2017deep,liu2020mnemonics} utilize synthesized exemplars by image synthesis models. The performance of these methods depend on the image synthesis models, so they require a large extra computation overhead to optimize the image synthesis models and might not be applicable in practical CIIC with a strict computation budget.
All of the above works focus on class-incremental image-level classification. Our work is the first work which considers the class-incremental video-level classification. In this work, we establish CIVC and propose to address the forgetting problem by video-level effective old-class knowledge transfer and efficient old-class exemplars keeping.

\section{Method}
\subsection{Video Classification}
Before delving into the details of CIVC, we first introduce the task of video classification. Given a category set $L$, we have a video dataset $D=\{(V_i,y_i),y_i\in L\}$ and each video sample is composed of a set of frames $V_i = \{v_{i,t}\}^{|V_i|}_{t=1}$. 
A video classification model $F(;\Theta)$ with parameters $\Theta$ is composed of a feature extractor $f(;\theta)$ and a classifier with parameters $\phi$, where $\Theta = \{\theta, \phi\}$. Given a video sample $V$, the feature map of the video sample $f(V;\theta)$ is a $C\times T \times H\times W$ tensor where $C$ is the number of channels and $T,H,W$ are the temporal and spatial dimensions. Let tensor $x_{t} \in \mathbb{R}^{C\times H\times W}$ denotes the feature map of frame $v_{t}$ in the video $V$.
$\Theta$ is optimized on $D$ by the following loss function:
\begin{equation}\label{eq_1}
\begin{aligned}
\mathcal{L}(\Theta) = \mathcal{L}_{CE}(\Theta),
\end{aligned}
\end{equation}
where $\mathcal{L}_{CE}$ is a cross-entropy loss.

\subsection{Class-Incremental Video Classification}\label{section3.2}
We define CIVC as follows. Assume the datasets $D^1, D^2, \cdots,$ in a data stream arrive in order and $D^k=\{(V^t_i,y^k_i),y^t_i\in L^k\}$ arrives at the $k$-th training session, where $\forall s, q, s\neq q \Rightarrow L^s \cap L^q = \varnothing$. At the first training session, $D^1$ arrives (considered as the base dataset) and $\Theta^1$ is trained from scratch on $D^1$ by Equ.~\eqref{eq_1}. 
Then as $D^k$ appears at the $k$-th session ($k>1$), our goal is to obtain a model $F(;\Theta^k)$ which recognizes all encountered classes in $L^1, L^2,\cdots,L^k$ well.

To achieve this goal, a simple way is keeping all the encounter datasets $D^1, D^2, \cdots, D^k$ and training the model $F(;\Theta^k)$ from scratch on the mixed dataset at the $k$-th session. However, it is impractical because training the model from scratch repeatedly on the large-scale mixed video dataset requires huge computation and storage overhead.
Another way is directly fine-tuning the model $F(;\Theta^k)$ initialized by $\Theta^{k-1}$ on the dataset $D^k$, which can greatly reduce the overall training overhead. 
But this method leads to the performance degradation of the model $F(;\Theta^k)$ on previous classes ($ie$, catastrophic forgetting). To meet the demand of real-world applications, we need to maintain the performance on previous video classes as well as reduce the overall overhead for training.

\begin{figure*}[t]
	\begin{center}
		\includegraphics[width=0.85\linewidth]{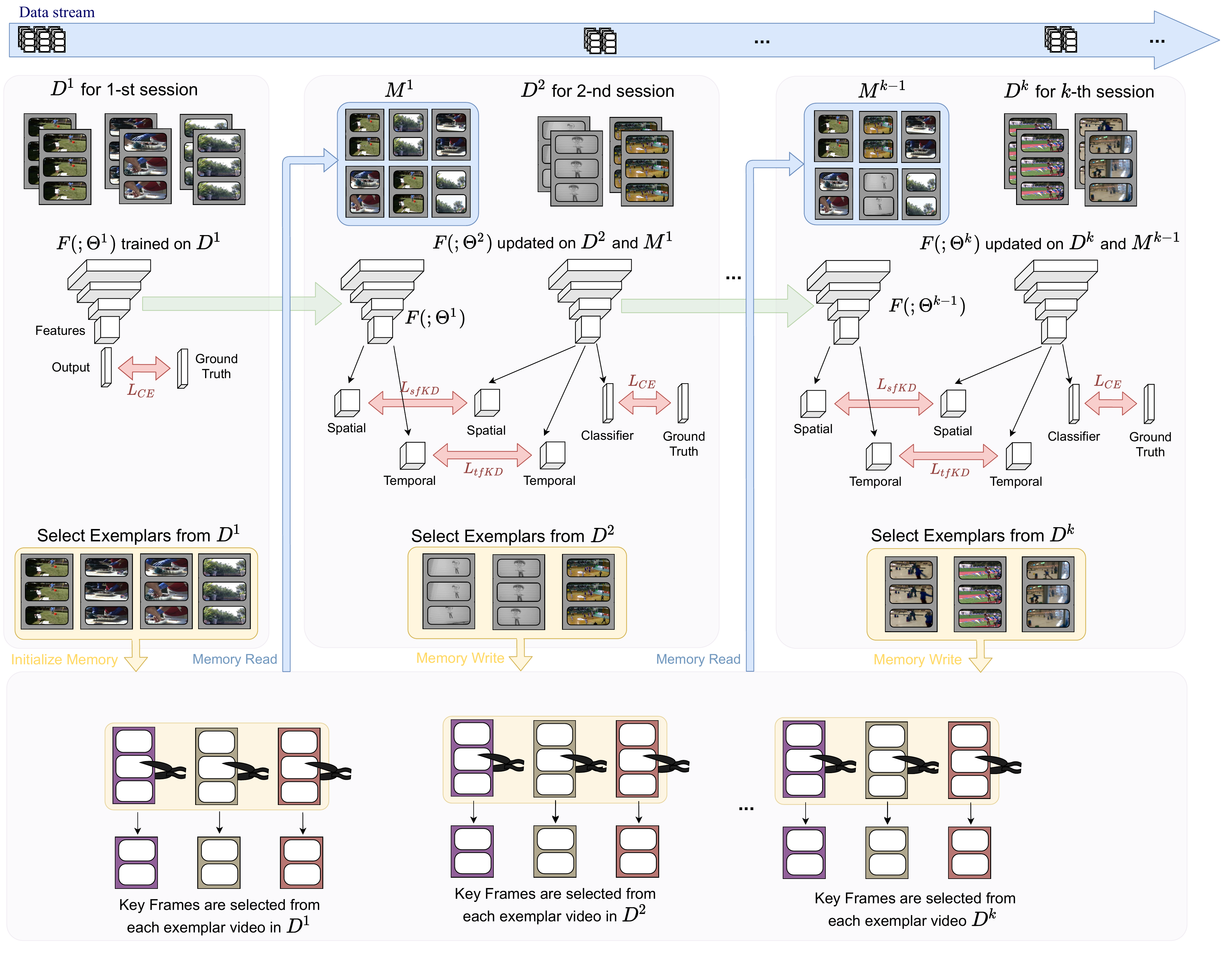}
	\end{center}
	\caption{Illustration of our CIVC framework. At each training session, the decomposed spatio-temporal knowledge transfer and the dual granularity exemplar selection are performed. For each session (except for the first session), the network from the precedent session serves as a teacher network and has supervisions on spatial feature and temporal feature of the new model by $\mathcal{L}_{sfKD}$ and $\mathcal{L}_{tfKD}$ respectively. The memory used to store key-frames of representative videos will be updated at the end of each session; the new arrived data $D^k$, combined with the memory $M^{k-1}$, is fed into the $k$-th session.}
	\label{fig:three}
	\vspace{-5pt}
\end{figure*}

Existing CIIC works maintain the performance on previous classes by knowledge distillation, and reduce the overall overhead by keeping only a few representative samples from the old datasets in a limited memory buffer $M$. Taking the $k$-th session as an example, these methods usually keep few exemplars $M^{k-1}$ drawn from previous datasets $D^1,\cdots,D^{k-1}$ and updating $\Theta^k$ by transferring knowledge from $\Theta^{k-1}$.
The approach for knowledge transfer and exemplar selection are introduced as follows.

\noindent\textbf{Knowledge Transfer.} To transfer the knowledge from the model $F(;\Theta^{k-1})$, the distillation loss term $\mathcal{L}_{KD}$ is added to the loss function while optimizing the model $F(;\Theta^k)$:
\begin{equation}
\label{eq_2}
\begin{aligned}
\mathcal{L}(\Theta^k) = \mathcal{L}_{CE}(\Theta^k)+ \gamma \mathcal{L}_{KD}(\Theta^k),
\end{aligned}
\end{equation} 
where $\gamma > 0$ is a hyper-parameter balancing the importance of the loss terms. The distillation loss is utilized to maintain the performance on previous classes and conducted in both the feature extractor and the classifier:
\begin{equation}
\label{eq_3}
\begin{aligned}
\mathcal{L}_{KD}(\Theta^k) = \mathcal{L}_{fKD}(\theta^k)+ \alpha \mathcal{L}_{cKD}(\Theta^k),
\end{aligned}
\end{equation} 
where $\mathcal{L}_{fKD}$ and $\mathcal{L}_{cKD}$ is the feature distillation loss and classifier distillation loss respectively. The implementation for $\mathcal{L}_{fKD}$ takes the form:
\begin{equation}
\label{eq_4}
\begin{aligned}
\mathcal{L}_{fKD}(\theta^k)=\sum_{(V,y)\in D^k \cup M^{k-1}}\left\|f(V;\theta^{k-1})-f(V;\theta^{k})\right\|^2,
\end{aligned}
\end{equation}
where $\left\|\cdot - \cdot \right\|$ denotes the Euclidean distance.
As for $\mathcal{L}_{cKD}(\Theta^k)$, a general form of implementation can be formulated as:
\begin{equation}
\begin{aligned}
\mathcal{L}_{cKD}(\Theta^k) &= \sum_{(V,y)\in D^k \cup M^{k-1}} \sum_{s=1}^{n}-\delta_s(V;\Theta^{k-1})\log(\delta_s(V;\Theta^{k})),\\
\delta_s(V;\Theta^k)&=\frac{e^{F_s(V;\Theta^{k})/TP}}{\sum_{j=1}^{n}e^{F_j(V;\Theta^{k})/TP}},\\
\end{aligned}
\end{equation}
where $n=\sum_{i=1}^{k-1}|L^{i}|$ is the number of the old classes and $TP$ is the distillation temperature.

\noindent\textbf{Exemplar Selection.} After updating the model $F(;\Theta^k)$, a common implementation first utilizes the feature extractor $f(;\theta^k)$ to extract the feature for each video $V^k_i$ in $D^k$.
For each class $c^k_m$ in $L^k$, the class center $U_{c^k_m}$ is obtained by computing the mean of all samples' features. Then the samples of each class $c^k_m$ are sorted according to their distances to the class center in ascending order. As a result, a class-specific ranking list is obtained for each class $c^k_m$. The top $K$
instances of the ranking list are picked out as the exemplars and stored in the memory $M^k$.

Currently, both the knowledge transfer method and the exemplar selection method are tailored for images. Compared with a single image, a video is much more complicated with an additional temporal dimension. In this paper, we focus on designing an efficient exemplar selection scheme and an effective knowledge transfer scheme tailored for video. Specifically, we propose a decomposed spatio-temporal knowledge transfer scheme in Section~\ref{video_Tra} and a dual granularity exemplar selection scheme in Section~\ref{video_exem}.  
The illustration of our scheme for CIVC is shown in Figure~\ref{fig:three}.

\begin{figure}[t]
	\begin{center}
		\includegraphics[width=0.95\linewidth]{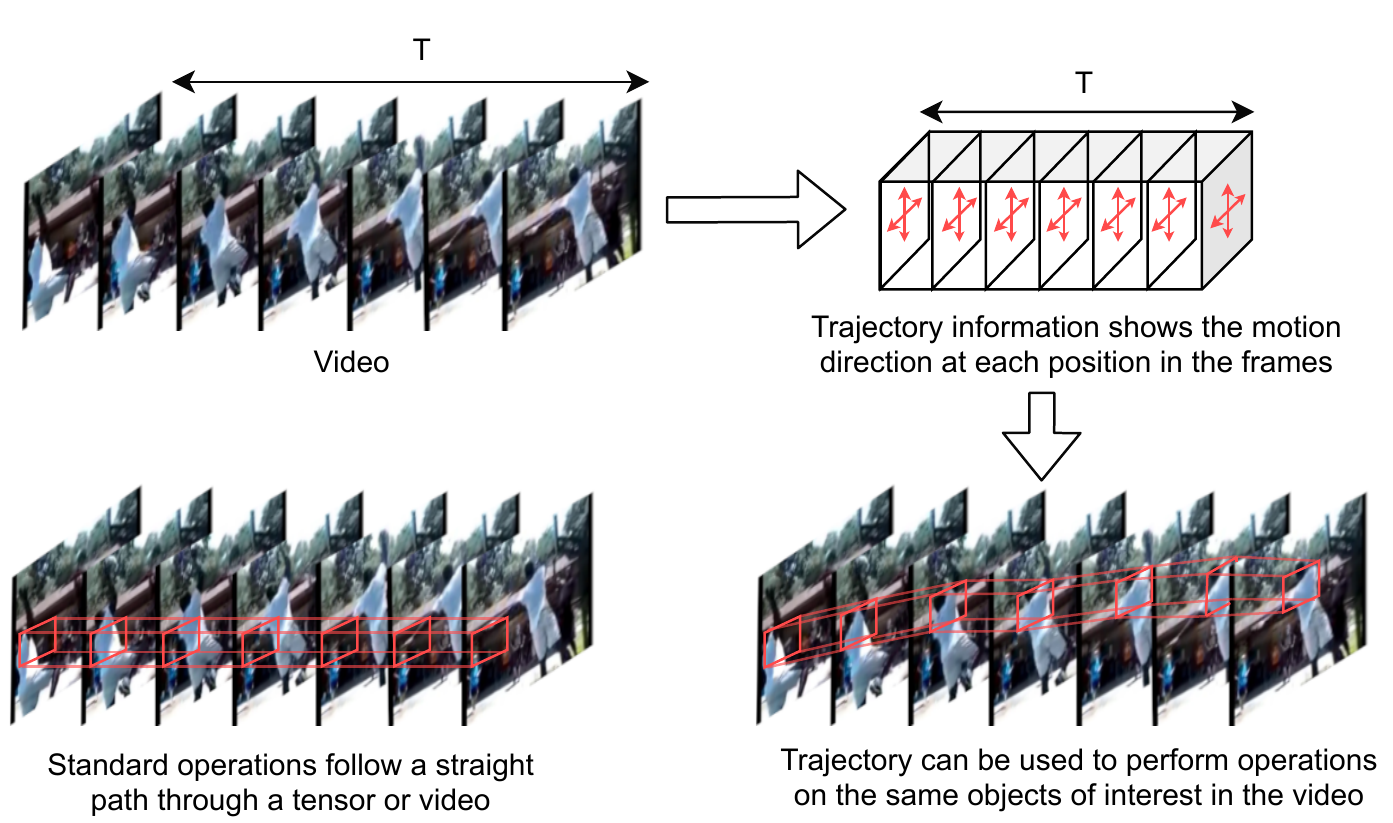}
	\end{center}
	\caption{Illustration of trajectory information. The trajectory information shows the motion direction at each position in the frames; hence, it can be used to better align the features contrast to the standard pooling operation}
	\label{fig:2-2}
	\vspace{-9pt}
\end{figure}
\subsection{Decomposed Spatio-Temporal Knowledge Transfer}\label{video_Tra}
To achieve effective knowledge transfer, we propose a video-tailored decomposed spatio-temporal knowledge transfer approach. We observe that directly distilling the fused spatio-temporal feature achieves less knowledge transfer than distilling the spatial and temporal features separately (as shown in Figure~\ref{fig:two}). 

In order to transfer the spatial knowledge and temporal knowledge separately, we first need to decompose the fused spatio-temporal feature of a video sample to a spatial feature and a temporal feature. For simplicity, we utilize $\Phi_{sf}(\cdot)$ and $\Phi_{tf}(\cdot)$ to denote the spatial and temporal feature decompose function respectively.
Then at the $k$-th session, the feature distillation loss in Equ.~\eqref{eq_3} is conducted for both the spatial feature and temporal feature:
\begin{equation}
\label{eq_6}
\begin{aligned}
\mathcal{L}_{fKD}(\theta^k)=\mathcal{L}_{sfKD}(\theta^k)+\lambda \mathcal{L}_{tfKD}(\theta^k),
\end{aligned}
\end{equation}
where $\mathcal{L}_{sfKD}$ and $\mathcal{L}_{tfKD}$ are the spatial feature and temporal feature distillation loss respectively, $\lambda$ is a hyper-parameter balancing the importance of the spatial and temporal feature distillation loss. $\mathcal{L}_{sfKD}$ and $\mathcal{L}_{tfKD}$ takes the similar form in Equ.~\eqref{eq_4}:
\begin{equation}
\begin{aligned}
\mathcal{L}_{sfKD}(\theta^k) = \sum_{(V,y)\in D^k \cup M^{k-1}}\left\|\Phi_{sf}(f(V;\theta^{k-1}))-\Phi_{sf}(f(V;\theta^{k}))\right\|^2,
\end{aligned}
\end{equation}

\begin{figure}[t]
	\begin{center}
		\includegraphics[width=0.8\linewidth]{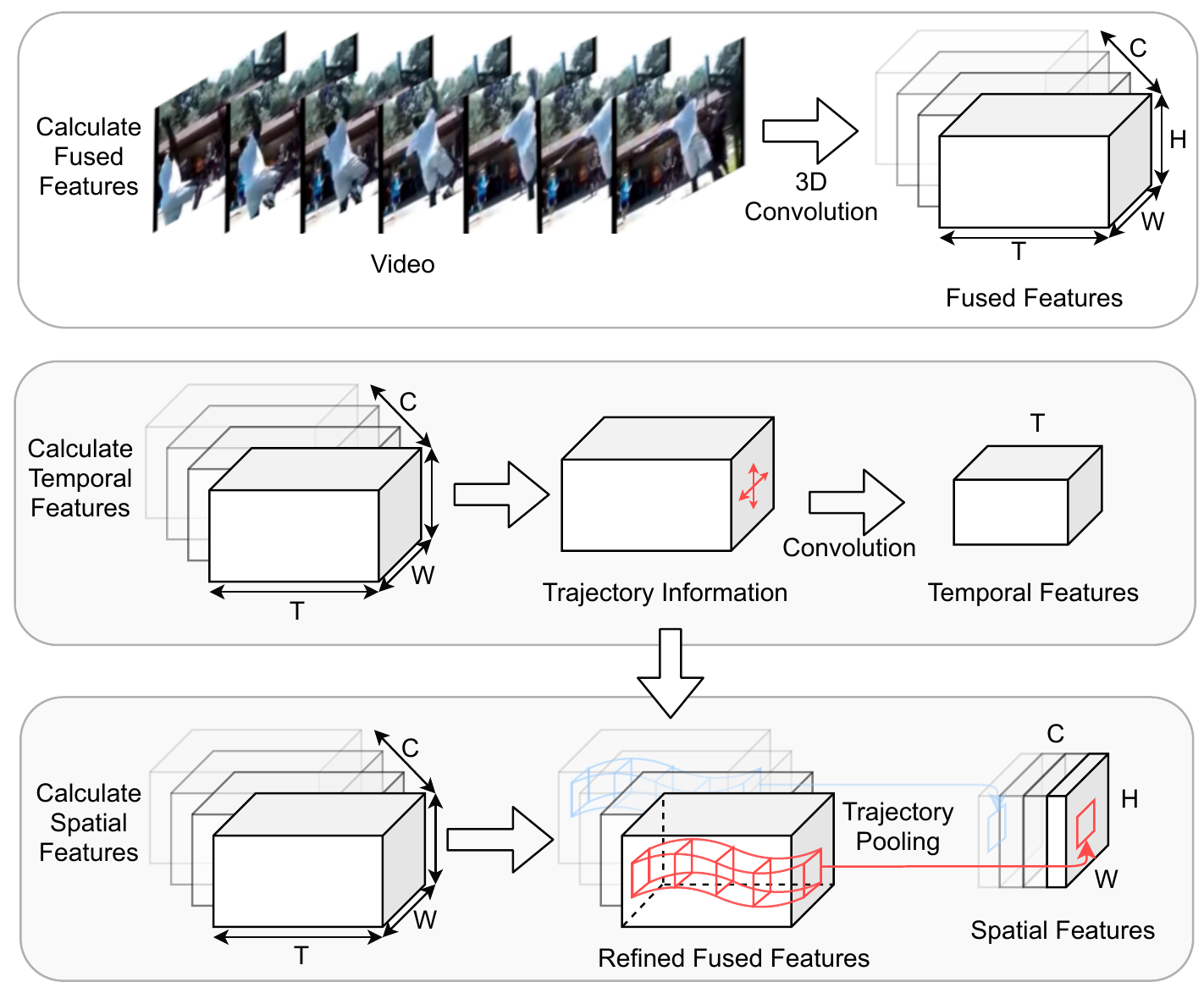}
	\end{center}
	\caption{Illustration of trajectory-based decomposed spatio-temporal knowledge transfer. The fused spatio-temporal feature is firstly obtained from a 3D CNN. The temporal feature is obtained from the trajectory information of the fused feature. The spatial feature is obtained from the refined fused feature aligned with the trajectory information.}
	\label{fig:2-3}
	\vspace{-8pt}
\end{figure}

Obviously, the choice of the temporal and spatial decompose function is essential for knowledge transfer. 
A simple implementation is to utilize the pooling operation to obtain the spatial feature from a fused spatio-temporal feature by pooling the temporal dimensions. For simplicity, we utilize $\mathtt{pool}_T(\cdot)$ to represent the pooling function at the temporal dimension ($\mathtt{pool}_W(\cdot)$ and $\mathtt{pool}_H(\cdot)$ for pooling at the width and height dimension) and then $\Phi_{sf}(f(V;\theta^{k})) = \mathtt{pool}_T(f(V;\theta^{k}))$. Similarly, the temporal feature is obtained from a fused spatio-temporal feature by pooling the spatial dimensions (pooling the width and height dimensions respectively), and then $\Phi_{tf}(f(V;\theta^{k}))$ is the concatenation of $\mathtt{pool}_W(f(V;\theta^{k}))$ and $\mathtt{pool}_H(f(V;\theta^{k}))$. 

However, the way of implementing $\Phi_{sf}(\cdot)$ by pooling is base on an implicit assumption that the fused spatio-temporal feature maps across frames are well aligned so that the features at the same locations (across consecutive frames) can be aggregated by pooling the temporal dimensions~\cite{zhao2018trajectory}. This assumption ignores the motion of people or objects and may lead to a distorted distillation. Some trajectory-based methods~\cite{wang2015action, zhao2018trajectory} are proposed to align the features across consecutive frames with the trajectory information (shown in Figure~\ref{fig:2-2}). Therefore, we also leverage the trajectory information while implementing $\Phi_{sf}(\cdot)$:
\begin{equation}
\begin{aligned}
\Phi_{sf}(f(V;\theta^{k})) = \mathtt{pool}_T(TrajAlign(f(V;\theta^{k})),
\end{aligned}
\end{equation}
where $TrajAlign(\cdot)$ denotes the trajectory-based feature alignment function. 
Specifically, the trajectory-based feature alignment function align the feature $x_t(p_t)$ of frame $v_t$ in $f(V;\theta^{k})$ at position $p_t=(h,w)\in [0,H) \times [0,W)$ by a 1D convolution operation (shown in Figure~\ref{fig:2-3}) and the convolution operation is conducted across irregular grids such that the sampled positions at different times correspond to the same physical point of a moving object:
\begin{equation}
\begin{aligned}
TrajAlign(x_t(p_t)) 
= \sum_{\tau=-\Delta t}^{\Delta t} {w_{\tau}x_{t+\tau}(\widetilde{p}_{t+\tau})},
\end{aligned}
\end{equation}
where $\{w_{\tau}:\tau\in[-\Delta t,\Delta t]\}$ are the parameters of the filter with kernel size $(2 \Delta t + 1)$.
The point $p_t$ at frame $v_t$ can be tracked to position $\widetilde{p}_{t+1}$ at next frame $v_{t+1}$ using the following equation:
\begin{equation}
\begin{aligned}
\widetilde{p}_{t+1}=p_t + \overrightarrow{o}(p_t).
\end{aligned}
\label{pt_pt1}
\end{equation}
where $\overrightarrow{o}(p_t)$ denotes a forward dense motion field of $p_t$~\cite{zhao2018trajectory}. For $\tau > 1$, the sample position $\widetilde{p}_{t+1}$ can be calculated by applying Equ~\eqref{pt_pt1} iteratively. To track the previous frame $v_{t-1}$, a backward dense motion field $\overleftarrow{o}(p_t)$ is used likewise. 

Moreover, the trajectory information itself can be considered as a temporal information pattern~\cite{wang2013action} and then we leverage the trajectory information while implementing $\Phi_{tf}(\cdot)$:
\begin{equation}
\begin{aligned}
\Phi_{tf}(f(V;\theta^{k})) = TrajTemp(f(V;\theta^{k})),
\end{aligned}
\end{equation}
where $TrajTemp(\cdot)$ denotes the trajectory-based temporal feature transformation function.
Specifically, the trajectory-based temporal feature transformation function describes the local temporal information pattern of $x_t(p_t)$ using the sequence of trajectory information in the form of motion field ($\ie$ $\overrightarrow{o}(p_t)$ $\overleftarrow{o}(p_t)$) at first, and then transform the trajectory information of $x_t$ ($\ie$ the concatenation of $\overrightarrow{o}(p_t)$ and $\overleftarrow{o}(p_t)$) to a temporal feature by convolution operation (shown in Figure~\ref{fig:2-3}).

\begin{figure}[t]
	\begin{center}
		\includegraphics[width=0.55\linewidth]{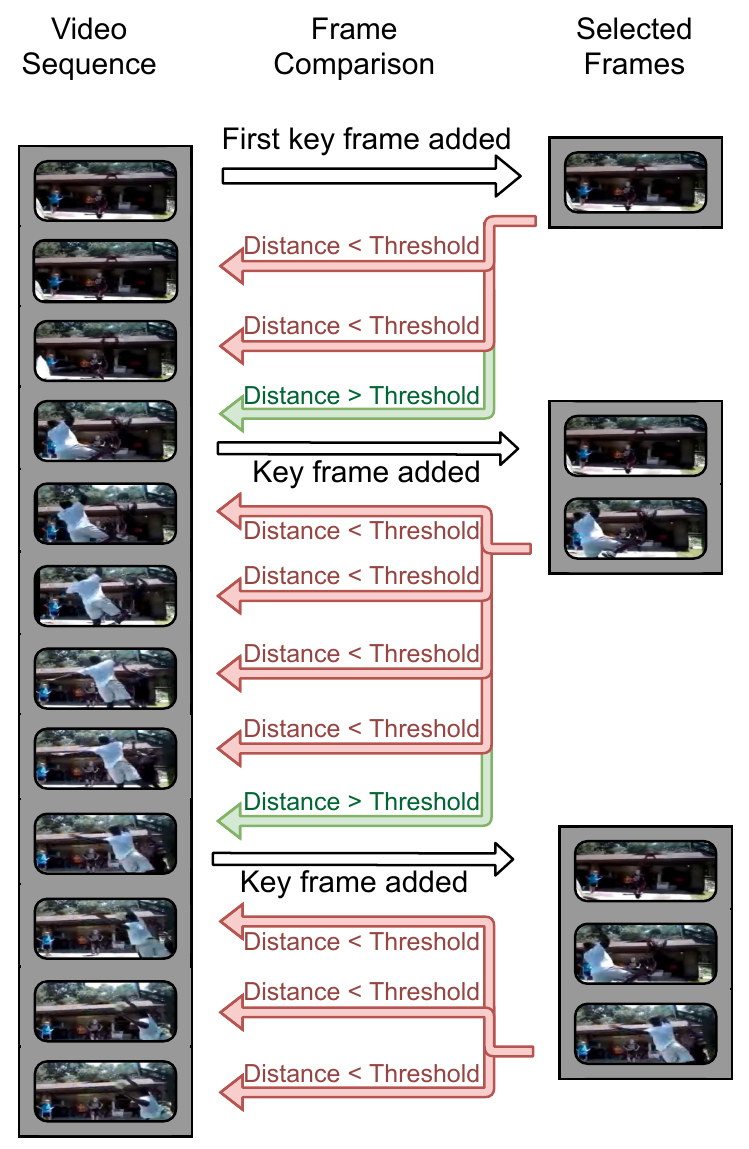}
	\end{center}
	\caption{Illustration of our inner-instance granularity key-frames selection. Each frame is compared with the newest key-frame. If the distance is larger than the threshold, the current frame will be added to key-frame set and marked as the newest key-frame; else this frame will be ignored.}
	\label{fig:3.4}
	\vspace{-11pt}
\end{figure}
\subsection{Dual Granularity Exemplar Selection}\label{video_exem}
In this section, we propose a video-tailored dual granularity exemplar selection approach to reduce the memory overhead. Existing CIIC methods focus on selecting representative samples at the instance granularity. However, being different from images, a video comprises a number of frames and each frame itself is an image. Therefore, we further investigate the representative at the inner-instance granularity. 

At each session, we first select the representative samples from the instance-granularity and the top $K$ samples are selected as the exemplars. As videos are comprised of a sequence of consistent frames, the information contained in neighbor frames may densely overlap and thus can be regarded as redundant information. The redundant information of multiple video frames can be reduced by keeping only a few key-frames. Suppose there is a video exemplar $V$, we select the key-frames by computing the distance between a newest key-frame $v_i$ and a frame $v_j$ ($i<j$). Specifically, the distance function is defined as follows:
\begin{equation}
\begin{aligned}
Dis(v_{i}, v_{j}) = \left\|v_{i}-v_{j}\right\|^2,
\end{aligned}
\end{equation}
If the distance between $v_{i}$ and $v_{j}$ is smaller than the threshold $Thre$, we keep the key-frame set unchanged; if the distance is larger than the threshold, the current frame is added to the key-frame set and the newest key-frame is now pointed to this frame, as illustrated in Figure~\ref{fig:3.4}:
\begin{equation}
\begin{aligned}
Thre = {\beta}_{thre} \sum_{i=1}^{|V|-1} Dis(v_i, v_{i+1}).
\end{aligned}
\end{equation} 
where ${\beta}_{thre}$ is a hyper-parameter which can affect the memory overhead for keeping exemplars.

\section{Experiments}
\begin{table*}[t]
	\centering
	\caption{Validation of decomposed spatio-temporal knowledge transfer scheme and dual granularity exemplar selection.}
	\resizebox{1\textwidth}{!}{
		\begin{tabular}{lccccccccccccccc}
			\toprule
			\multirow{2}{*}{Method}&  \multicolumn{3}{c}{i-Sth-Sth-B0}& & \multicolumn{3}{c}{i-Sth-Sth-B20} & &\multicolumn{3}{c}{i-Kinetics-B0}&&  \multicolumn{3}{c}{i-Kinetics-B20}\\
			\cline{2-4} \cline{6-8} \cline{10-12}\cline{14-16}
			&\emph{Acc.(\%)$\uparrow $}& \emph{Forget.(\%)$\downarrow $}& \emph{Mem.(G)$\downarrow $}&&\emph{Acc.} & \emph{Forget.}&\emph{Mem.}&&\emph{Acc.} & \emph{Forget.}&\emph{Mem.}&&\emph{Acc.} & \emph{Forget.}&\emph{Mem.} \\
			\midrule
			Baseline 						& 40.79&40.60&0.61 && 34.45 &43.15&0.61&&52.26&30.49&3.30&&48.97& 29.78 &3.30\\
			\midrule
			Baseline+decomposed-pool	    & 45.87&33.60&0.61 && 38.83 &36.20&0.61&&56.78&25.51&3.30&&53.53& 23.95 &3.30 \\
			Baseline+decomposed-traj        & \textbf{46.30}&\textbf{32.70}&0.61 && \textbf{40.04 }&\textbf{34.55}&0.61&&\textbf{57.65}&\textbf{24.73}&3.30&&\textbf{54.75}& \textbf{21.96} &3.30 \\
			\midrule
			Baseline+dual-gra        & 40.35 & 41.24 & \textbf{0.44} & & 34.08 & 43.46 & \textbf{0.44}  && 51.97 & 30.92 & \textbf{1.52} && 48.54 & 30.26 & \textbf{1.52}  \\
			\bottomrule
		\end{tabular}
	}
	\label{tab:four:three}
\end{table*}
\subsection{Datasets}
For the task of video classification, existing datasets can be roughly divided into two groups: YouTube-type videos and crowd-sourced videos~\cite{xie2018rethinking, zhou2018temporal}. Crowd-sourced videos usually focus more on modeling the temporal relationships, since visual contents among different classes are more similar than those of YouTube-type videos. We select two popular large-scale video classification datasets Something-Something V2~\cite{goyal2017something} ($\ie$ Sth-Sth) and Kinetics~\cite{kay2017kinetics} and conduct comprehensive experiments on them. Something-Something V2 is a crowd-sourced video dataset that includes 220k video clips for 174 fine-grained classes; Kinetics is a YouTube-type video dataset that contains 400 human action categories and provides 240k training videos and 20k validation videos. We utilize these two datasets to construct the CIVC benchmarks in Section~\ref{section4.2}. 

\subsection{Benchmark and Evaluation Protocol}\label{section4.2}
\noindent\textbf{Benchmark.} Four CIVC benchmarks are constructed on Something-Something V2 and Kinetics: 1) i-Something-Something-B0: selecting a subset of 40 classes from Something-Something V2, where all 40 classes are divided into 4 splits and trained in 4 training sessions. 2) i-Something-Something-B20: utilizing a subset of 40 classes from Sth-Sth V2, where the model is trained on 20 classes first and the remaining 20 classes are divided into 5 sessions. 3) i-Kinetics-B0: randomly selecting a subset of 40 classes from Kinetics, where all 40 classes are divided into 4 splits and trained in 4 sessions.
4) i-Kinetics-B20: selecting a subset of 40 classes from Kinetics randomly, where the model is first trained with 20 classes and the remaining 20 classes are divided into 5 training sessions.

\noindent\textbf{Evaluation protocol.} Each method is trained on the CIVC benchmark in several sessions. At the end of each session, we report the classification accuracy of the trained model $F(;\Theta^k)$ on the test part data $D^1_{test}, D^2_{test}, \cdots, D^k_{test},$ and the forgetting rate~\cite{yu2020semantic, chaudhry2018riemannian} by calculating the difference between the accuracies of $F(;\Theta^1)$ and $F(;\Theta^k)$ on the same initial test data $D^1_{test}$. After all sessions, we report the accuracy and the forgetting rate at the last session (denoted by ``\emph{Acc. (\%)}" and ``\emph{Forget. (\%)}") as the final evaluation. In addition, we report the memory overhead for keeping the exemplars, denoted as \emph{Mem. (G)}, which is usually constrained in the real-world application.

\subsection{Implementation Details}
We follow the video preprocessing procedure introduced in TSN~\cite{wang2016temporal}. For training, we first resize each frame in the video to $256\times256$ and then randomly crop a $224\times224$ region from the video clip. We sparsely and uniformly sample $8$ segments per video. For inference, we change the random crop to the center crop. On Kinetics dataset, we randomly apply horizontal flip during training. Since the label in Something-Something V2 dataset incorporates concepts of left and right, e.g., pulling something from left to right and pulling something from right to left, we do not use horizontal flip for this dataset. 
In this paper, we focus on maintaining good performance with limited resource. Thus, we choose TSM-ResNet50~\cite{lin2019tsm} which extracts 3D features to model the spatial and temporal information as our backbone.
We initialize the network using pre-trained weights on ImageNet~\cite{deng2009imagenet}. We optimize our model with SGD~\cite{bottou2010large}. At the first training session, the network is trained from an initial learning rate of $0.005$ and is reduced by $0.1$ after $35$ and $45$ epochs. The whole training procedure takes $50$ epochs. And for the remaining sessions, the network is trained from an initial learning rate of $0.0025$ and is reduced by $0.1$ after $17$ and $22$ epochs. The whole training procedure takes $25$ epochs. The loss weights $\alpha$, $\gamma$, $\lambda$ are all set to $1$. The distillation temperature $TP$ is set to $2$, and the number of exemplars per class is set to $10$ unless otherwise stated. For a fair comparison, the training setting and backbone are the same for all methods. 
The hyper-parameter ${\beta}_{thre}$ on i-Sth-Sth and i-Kinetics is set to $0.9$ and $1.05$ respectively. 
The motion field ${o}$ is often represented by dense optical flow in previous works~\cite{DBLP:conf/cvpr/WangKSL11, wang2013action, wang2015action}. Since calculating optical flow is too time-consuming, we integrate a learnable module to quickly calculate the approximate optical flow~\cite{kwon2020motionsqueeze}.
We implemented our framework with PyTorch~\cite{paszke2017automatic} and conduct experiments with four Nvidia 3060 GPUs.

\subsection{Ablation Study}
In this section, we first carry out ablation experiments to validate the effectiveness of our decomposed spatio-temporal knowledge transfer scheme, and compare different implementations of it. Then we conduct experiments to validate our dual granularity exemplar selection scheme. All of them are conducted on four benchmarks.

\noindent\textbf{Baseline.} Our main baseline is given based on a classical CIIC method EEIL~\cite{castro2018end}, which selects exemplars of previously seen classes only from the instance-granularity and maintains the performance on previous classes by the classifier distillation loss. All experiments are reported with our implementation unless specified otherwise. The original EEIL fixes the total number of exemplars stored at any session, and changes the number of exemplars per class depending on the total number of classes. Unlike the original EEIL, we fix the number of exemplars per class as in our implementation. We extend the original implementation of EEIL by applying a fused spatio-temporal feature distillation loss as the variant has been shown to improve the accuracy. We refer to the resulting variant as our ``Baseline".

\begin{table*}[t]
	\centering
	\caption{Accuracy, forgetting rate and memory overhead of different methods on the four CIVC benchmarks.}
	\resizebox{1\textwidth}{!}{
		\begin{tabular}{lccccccccccccccc}
			\toprule
			\multirow{2}{*}{Method}&  \multicolumn{3}{c}{i-Sth-Sth-B0}  && \multicolumn{3}{c}{i-Sth-Sth-B20}& &\multicolumn{3}{c}{i-Kinetics-B0} & & \multicolumn{3}{c}{i-Kinetics-B20}\\
			\cline{2-4} \cline{6-8}\cline{10-12} \cline{14-16}
			&\emph{Acc.(\%)$\uparrow $}& \emph{Forget.(\%)$\downarrow $}& \emph{Mem.(G)$\downarrow $}& &\emph{Acc.} & \emph{Forget.}& \emph{Mem.}&&\emph{Acc.} & \emph{Forget.}& \emph{Mem.}&&\emph{Acc.} & \emph{Forget.}& \emph{Mem.}\\
			\midrule
			FT-Lower Bound									 & 11.52 & 78.20 & 0.00&  &  7.62 & 71.00 & 0.00& & 21.25 & 83.47 & 0.00 && 8.37 & 82.83 &  0.00\\
			\midrule
			\midrule
			iCaRL~\cite{rebuffi2017icarl}        			 & 34.37 & 49.53 & 0.61&  & 29.92 & 47.21 & 0.61& & 47.64 & 41.63 & 3.30 & & 41.89 & 36.98 & 3.30\\
			EEIL~\cite{castro2018end} 			 			 & 39.43 & 47.62 & 0.61&  & 31.60 & 44.64 & 0.61& & 51.43 & 36.47  &3.30 & & 47.79 & 33.34 & 3.30 \\
			BiC~\cite{wu2019large} 							 & 41.70 & 42.50 & 0.61&  & 33.24 & 44.00 & 0.61& & 54.62 & 34.29  &3.30 & & 50.26 & 28.15 & 3.30 \\
			PODNet~\cite{douillard2020podnet}    			 & 42.07 & 37.36 & 0.61&  & 35.98& 40.62 & 0.61& & 53.90 &  28.98 &3.30 & & 51.03&24.90 & 3.30\\
			\midrule
			Ours (ST)								 		 & 46.03 & 31.72 & \textbf{0.44}&  & 39.73 & 34.75 & \textbf{0.44}& & 57.26& 25.50  & \textbf{1.52} & & 54.24 & 22.90 & \textbf{1.52}\\
			Ours (SM)								 		 & \textbf{48.17} & \textbf{30.72} & 0.61&  & \textbf{45.84} & \textbf{32.25} & 0.61& & \textbf{60.21} & \textbf{19.59}& 3.30 &  & \textbf{60.41} & \textbf{18.90}  & 3.30\\
			\midrule
			\midrule
			Joint-Upper Bound                                & 64.65 &  - & 61.04 & & 64.65 & - & 61.04 && 72.95 & - &248.81&  & 72.95 & - & 248.81 \\
			\bottomrule
		\end{tabular}
	}
	\label{tab:one}
\end{table*}

\begin{figure*}[t]
	\begin{minipage}[ht]{1\textwidth}
		\centering
		\includegraphics[width =0.245\columnwidth]{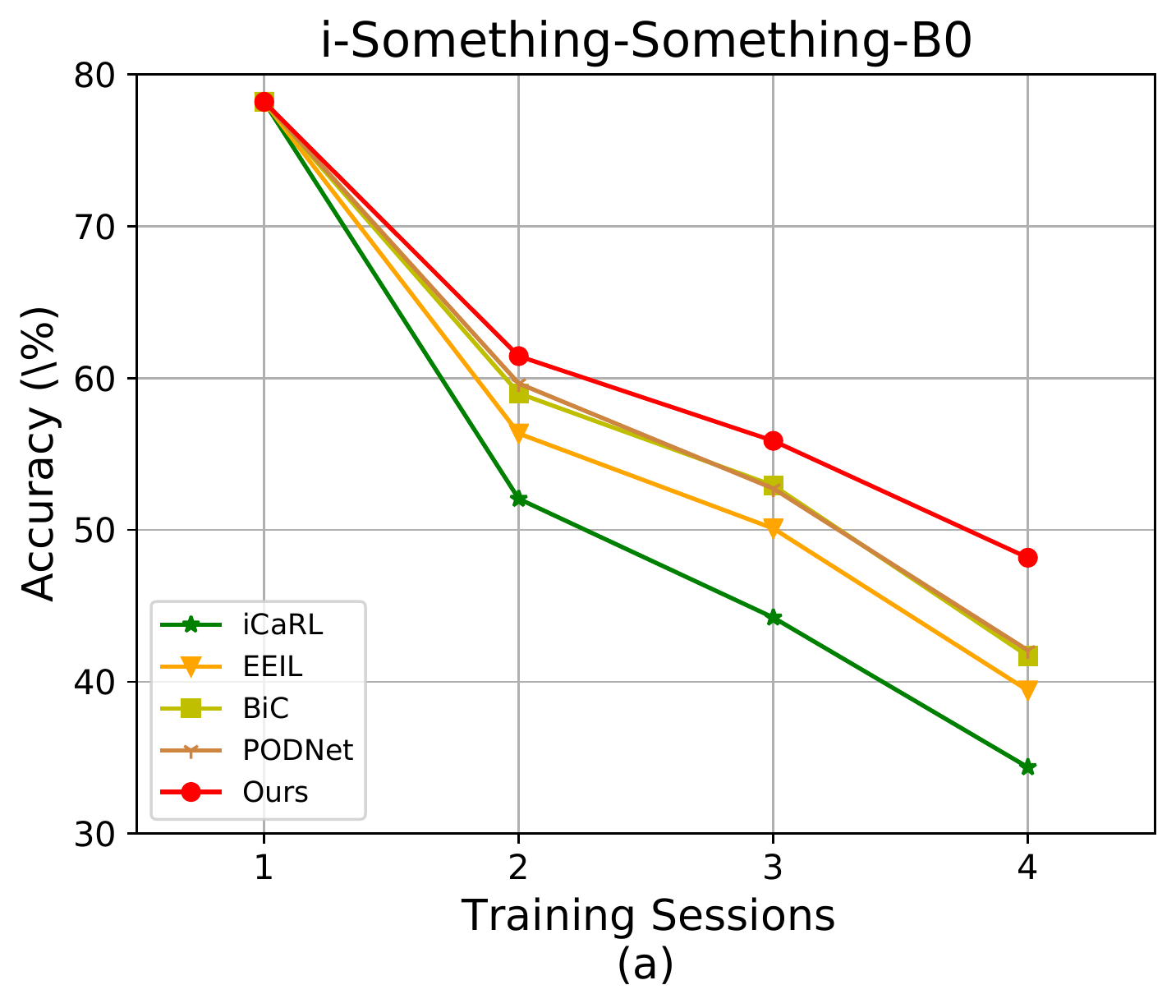}	
		\includegraphics[width =0.245\columnwidth]{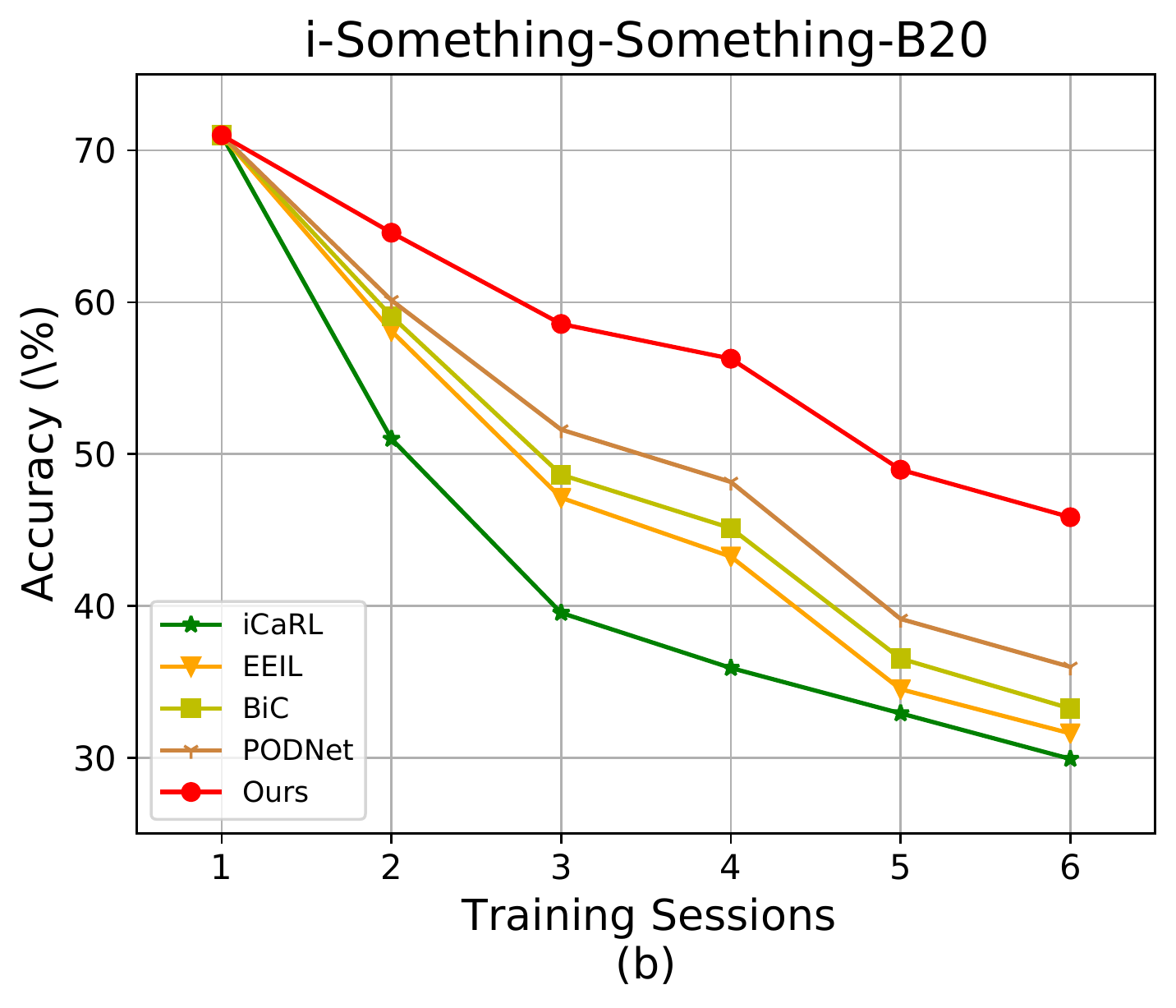}	
		\includegraphics[width = 0.245\columnwidth]{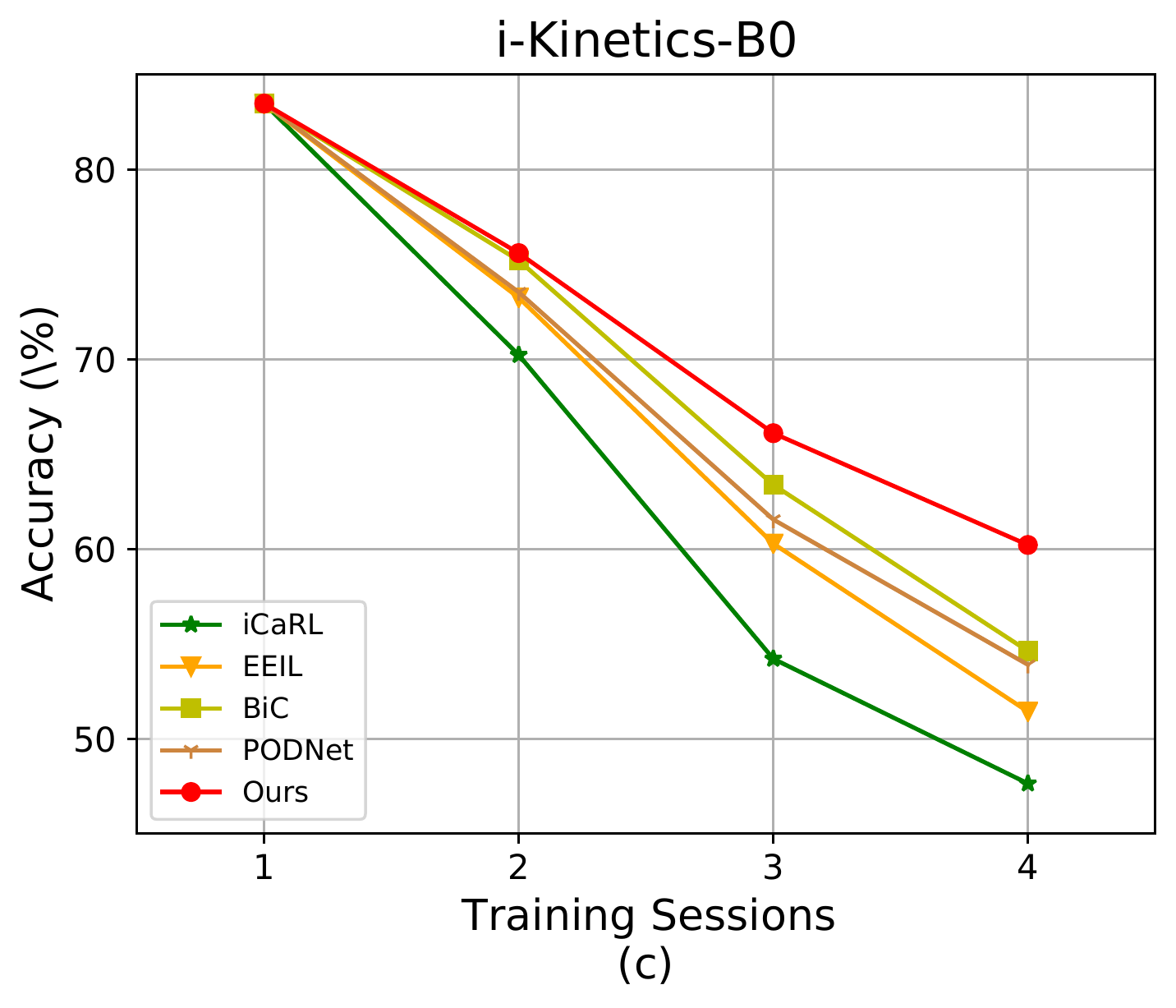}
		\includegraphics[width = 0.245\columnwidth]{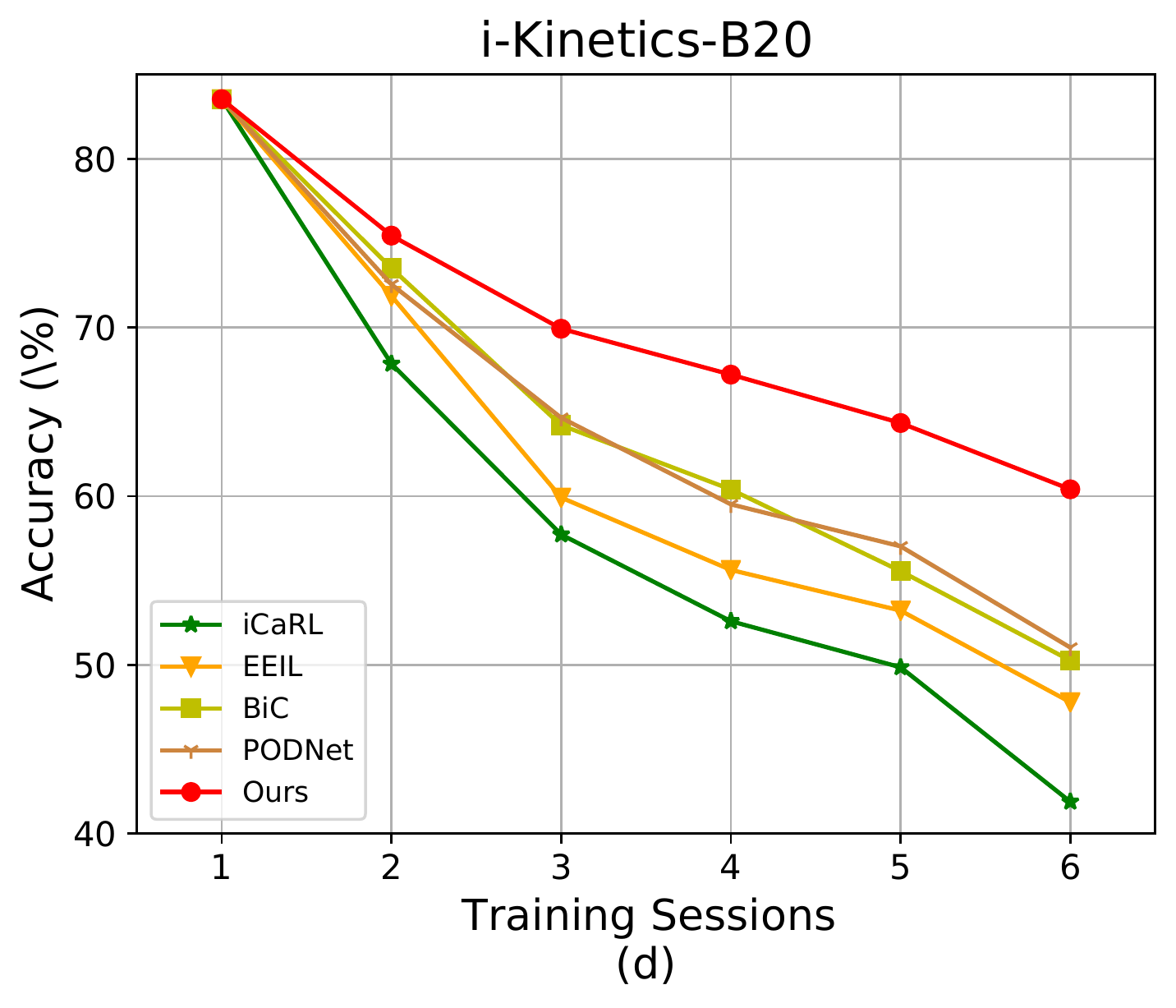}
	\end{minipage}	
	\caption{The accuracy curves of different methods on the four CIVC benchmarks.}
	\label{fig:exp:one}
\end{figure*}

\noindent\textbf{Effect of our decomposed spatio-temporal knowledge transfer scheme.}
In order to demonstrate the effectiveness of our decomposed spatio-temporal Knowledge transfer scheme, we compare the performance of ``Baseline" with our decomposed spatio-temporal knowledge transfer scheme (denoted by ``Baseline + decomposed''). The curves of accuracy are shown in Figure~\ref{fig:two}. We can see that the accuracy of ``Baseline+decompose" surpasses that of ``Baseline" at every session for different benchmarks. Moreover, the gap between ``Baseline + decomposed" and ``Baseline" increases with continuous adding of novel classes. In particular, the accuracy of ``Baseline + decompose" outperforms ``Baseline" by around $6\%$ at the last session on i-Kinetics-B20. This result indicates that distilling the spatial and temporal feature separately achieves more effective knowledge transfer than directly distilling the fused spatio-temporal feature.

\noindent\textbf{Different decomposed spatio-temporal knowledge transfer implementations} We evaluate the decomposed spatio-temporal knowledge transfer scheme with a simple pooling-based implementation (denoted by ``Baseline + decomposed-pool") and a complex trajectory-based implementation (denoted by ``Baseline + decomposed-traj"). We compare the performance of ``Baseline + decomposed-traj" and ``Baseline + decomposed-pool" and the results are shown in Table~\ref{tab:four:three}. It is observed that the accuracy of Baseline + decomposed-traj outperforms that of Baseline + decomposed-pool for all benchmarks. Moreover, Baseline + decomposed-traj achieves less forgetting rate than Baseline + decomposed-pool.

\noindent\textbf{Effect of our dual granularity exemplar selection scheme.}
We evaluate the performance of ``Baseline" with our dual granularity exemplar selection scheme (denoted by ``Baseline + dual-gra") and Table~\ref{tab:four:three} summarizes the experimental results for different benchmarks. We can see that ``Baseline + dual-gra" achieves similar performance with fewer memory overhead than ``Baseline". In particular, ``Baseline + dual-gra" achieves similar accuracy than ``Baseline" with only half of the memory overhead on i-Kinetics-B0. This result indicates that the inner-instance granularity key-frames selection is essential to reduce the memory overhead for CIVC.

\subsection{Comparison to State-of-the-Art Methods}
In this section, we evaluate the CIVC performance of our proposed method on the four benchmarks, against other standard CIIC methods on the considered CIVC task, including iCARL~\cite{rebuffi2017icarl}, EEIL~\cite{castro2018end}, BiC~\cite{wu2019large} and PODNet~\cite{douillard2020podnet}. In the tables, we also report other two methods: simple fine-tuning $F(;\Theta^k)$ on $D^k$ (denoted by ``FT") and training the model on all classes off-line (denoted by ``Joint"). The former can be considered as a lower bound and the latter can be regarded as an upper bound. 

Table~\ref{tab:one} summarizes the experimental results for the four CIVC benchmarks. We can see that the average accuracy of our method consistently outperforms that of other methods by a sizable margin on different benchmarks. As for the memory overhead, ``Joint" keeps all previous datasets in memory and thus utilizes the largest memory overhead. iCaRL, EEIL and PODNet all select a fixed number of video exemplars per class from the instance-granularity and they utilize the same size of memory overhead. Our method first selects the same number of video exemplars and then keeps a small number of key-frames in memory from dual granularity and achieves lower memory overhead (denoted by ``Ours (ST)"). It is worth noting that our method with the same size of memory overhead (denoted by ``Ours (SM)") can further improve the performance. The overview on forgetting rates reveals that our approach is greatly helpful to maintain the performance on previous classes. For example, the forgetting rate of ``Ours (SM)" is $6.64\%$ less than that of PODNet on i-Sth-Sth-B0 ($8.37\%$ on i-Sth-Sth-B20, $9.39\%$ on i-Kinetics-B0, $6.00\%$ on i-Kinetics-B20). 
The accuracy curves of these methods are shown in Figure~\ref{fig:exp:one}. We can see that our method consistently surpasses the other methods for all training sessions on each benchmark. As the number of training sessions increases, it is observed that the margin between our method with other methods continuously increases, which indicates our method performs better on the benchmark with more number of training sessions.

\section{Conclusion}
In this paper, we propose the task of Class-Incremental Video Classification and design a novel framework for it. 
Our model inherits knowledge from old class videos effectively with the decomposed spatial-temporal knowledge transfer mechanism. 
By selecting both the video exemplars and key-frames inside the videos, we greatly reduce the storage memory and preserve the most representative information of old classes. Lastly, we establish the benchmark for this task and perform intensive evaluations of our method and other SOTA class-incremental learning methods, showing the effectiveness of our method.

\begin{acks}	
This work is supported in part by National Key Research and Development Program of China under Grant 2020AAA0107400, Zhejiang Provincial Natural Science Foundation of China under Grant LR19F020004, key scientific technological innovation research project by Ministry of Education, National Natural Science Foundation of China under Grant U20A20222, and Zhejiang Lab.
\end{acks}

\newpage
\bibliographystyle{ACM-Reference-Format}
\bibliography{sample-base}


\begin{thebibliography}{66}


\ifx \showCODEN    \undefined \def \showCODEN     #1{\unskip}     \fi
\ifx \showDOI      \undefined \def \showDOI       #1{#1}\fi
\ifx \showISBNx    \undefined \def \showISBNx     #1{\unskip}     \fi
\ifx \showISBNxiii \undefined \def \showISBNxiii  #1{\unskip}     \fi
\ifx \showISSN     \undefined \def \showISSN      #1{\unskip}     \fi
\ifx \showLCCN     \undefined \def \showLCCN      #1{\unskip}     \fi
\ifx \shownote     \undefined \def \shownote      #1{#1}          \fi
\ifx \showarticletitle \undefined \def \showarticletitle #1{#1}   \fi
\ifx \showURL      \undefined \def \showURL       {\relax}        \fi
\providecommand\bibfield[2]{#2}
\providecommand\bibinfo[2]{#2}
\providecommand\natexlab[1]{#1}
\providecommand\showeprint[2][]{arXiv:#2}

\bibitem[\protect\citeauthoryear{Ballas, Delezoide, and Pr\^{e}teux}{Ballas
  et~al\mbox{.}}{2012}]%
        {10.1145/2393347.2396511}
\bibfield{author}{\bibinfo{person}{Nicolas Ballas}, \bibinfo{person}{Bertrand
  Delezoide}, {and} \bibinfo{person}{Fran\c{c}oise Pr\^{e}teux}.}
  \bibinfo{year}{2012}\natexlab{}.
\newblock \showarticletitle{Trajectory Signature for Action Recognition in
  Video}. In \bibinfo{booktitle}{\emph{Proceedings of the 20th ACM
  International Conference on Multimedia}}. \bibinfo{publisher}{Association for
  Computing Machinery}, \bibinfo{pages}{1429–1432}.
\newblock


\bibitem[\protect\citeauthoryear{Bottou}{Bottou}{2010}]%
        {bottou2010large}
\bibfield{author}{\bibinfo{person}{L{\'e}on Bottou}.}
  \bibinfo{year}{2010}\natexlab{}.
\newblock \showarticletitle{Large-scale machine learning with stochastic
  gradient descent}.
\newblock In \bibinfo{booktitle}{\emph{Proceedings of COMPSTAT'2010}}.
  \bibinfo{publisher}{Springer}, \bibinfo{pages}{177--186}.
\newblock


\bibitem[\protect\citeauthoryear{Cao, Ji, Cao, Chang, and Niebles}{Cao
  et~al\mbox{.}}{2020}]%
        {cao2020few}
\bibfield{author}{\bibinfo{person}{Kaidi Cao}, \bibinfo{person}{Jingwei Ji},
  \bibinfo{person}{Zhangjie Cao}, \bibinfo{person}{Chien-Yi Chang}, {and}
  \bibinfo{person}{Juan~Carlos Niebles}.} \bibinfo{year}{2020}\natexlab{}.
\newblock \showarticletitle{Few-shot video classification via temporal
  alignment}. In \bibinfo{booktitle}{\emph{Proceedings of the IEEE/CVF
  Conference on Computer Vision and Pattern Recognition}}.
  \bibinfo{pages}{10618--10627}.
\newblock


\bibitem[\protect\citeauthoryear{Carreira and Zisserman}{Carreira and
  Zisserman}{2017}]%
        {carreira2017quo}
\bibfield{author}{\bibinfo{person}{Joao Carreira} {and} \bibinfo{person}{Andrew
  Zisserman}.} \bibinfo{year}{2017}\natexlab{}.
\newblock \showarticletitle{Quo vadis, action recognition? a new model and the
  kinetics dataset}. In \bibinfo{booktitle}{\emph{proceedings of the IEEE
  Conference on Computer Vision and Pattern Recognition}}.
  \bibinfo{pages}{6299--6308}.
\newblock


\bibitem[\protect\citeauthoryear{Castro, Mar{\'\i}n-Jim{\'e}nez, Guil, Schmid,
  and Alahari}{Castro et~al\mbox{.}}{2018}]%
        {castro2018end}
\bibfield{author}{\bibinfo{person}{Francisco~M Castro},
  \bibinfo{person}{Manuel~J Mar{\'\i}n-Jim{\'e}nez},
  \bibinfo{person}{Nicol{\'a}s Guil}, \bibinfo{person}{Cordelia Schmid}, {and}
  \bibinfo{person}{Karteek Alahari}.} \bibinfo{year}{2018}\natexlab{}.
\newblock \showarticletitle{End-to-end incremental learning}. In
  \bibinfo{booktitle}{\emph{Proceedings of the European conference on computer
  vision (ECCV)}}. \bibinfo{pages}{233--248}.
\newblock


\bibitem[\protect\citeauthoryear{Chaudhry, Dokania, Ajanthan, and
  Torr}{Chaudhry et~al\mbox{.}}{2018}]%
        {chaudhry2018riemannian}
\bibfield{author}{\bibinfo{person}{Arslan Chaudhry}, \bibinfo{person}{Puneet~K
  Dokania}, \bibinfo{person}{Thalaiyasingam Ajanthan}, {and}
  \bibinfo{person}{Philip~HS Torr}.} \bibinfo{year}{2018}\natexlab{}.
\newblock \showarticletitle{Riemannian walk for incremental learning:
  Understanding forgetting and intransigence}. In
  \bibinfo{booktitle}{\emph{Proceedings of the European Conference on Computer
  Vision (ECCV)}}. \bibinfo{pages}{532--547}.
\newblock


\bibitem[\protect\citeauthoryear{Delange, Aljundi, Masana, Parisot, Jia,
  Leonardis, Slabaugh, and Tuytelaars}{Delange et~al\mbox{.}}{2021}]%
        {delange2021continual}
\bibfield{author}{\bibinfo{person}{Matthias Delange}, \bibinfo{person}{Rahaf
  Aljundi}, \bibinfo{person}{Marc Masana}, \bibinfo{person}{Sarah Parisot},
  \bibinfo{person}{Xu Jia}, \bibinfo{person}{Ales Leonardis},
  \bibinfo{person}{Greg Slabaugh}, {and} \bibinfo{person}{Tinne Tuytelaars}.}
  \bibinfo{year}{2021}\natexlab{}.
\newblock \showarticletitle{A continual learning survey: Defying forgetting in
  classification tasks}.
\newblock \bibinfo{journal}{\emph{IEEE Transactions on Pattern Analysis and
  Machine Intelligence}} (\bibinfo{year}{2021}).
\newblock


\bibitem[\protect\citeauthoryear{Deng, Dong, Socher, Li, Li, and Fei-Fei}{Deng
  et~al\mbox{.}}{2009}]%
        {deng2009imagenet}
\bibfield{author}{\bibinfo{person}{Jia Deng}, \bibinfo{person}{Wei Dong},
  \bibinfo{person}{Richard Socher}, \bibinfo{person}{Li-Jia Li},
  \bibinfo{person}{Kai Li}, {and} \bibinfo{person}{Li Fei-Fei}.}
  \bibinfo{year}{2009}\natexlab{}.
\newblock \showarticletitle{Imagenet: A large-scale hierarchical image
  database}. In \bibinfo{booktitle}{\emph{Proceedings of the IEEE conference on
  computer vision and pattern recognition (CVPR)}}.
\newblock


\bibitem[\protect\citeauthoryear{Douillard, Cord, Ollion, Robert, and
  Valle}{Douillard et~al\mbox{.}}{2020}]%
        {douillard2020podnet}
\bibfield{author}{\bibinfo{person}{Arthur Douillard}, \bibinfo{person}{Matthieu
  Cord}, \bibinfo{person}{Charles Ollion}, \bibinfo{person}{Thomas Robert},
  {and} \bibinfo{person}{Eduardo Valle}.} \bibinfo{year}{2020}\natexlab{}.
\newblock \showarticletitle{Podnet: Pooled outputs distillation for small-tasks
  incremental learning}. In \bibinfo{booktitle}{\emph{Computer vision-ECCV
  2020-16th European conference, Glasgow, UK, August 23-28, 2020, Proceedings,
  Part XX}}, Vol.~\bibinfo{volume}{12365}. Springer, \bibinfo{pages}{86--102}.
\newblock


\bibitem[\protect\citeauthoryear{Feichtenhofer}{Feichtenhofer}{2020}]%
        {DBLP:conf/cvpr/Feichtenhofer20}
\bibfield{author}{\bibinfo{person}{Christoph Feichtenhofer}.}
  \bibinfo{year}{2020}\natexlab{}.
\newblock \showarticletitle{{X3D:} Expanding Architectures for Efficient Video
  Recognition}. In \bibinfo{booktitle}{\emph{{IEEE/CVF} Conference on Computer
  Vision and Pattern Recognition, {CVPR}}}. \bibinfo{publisher}{{IEEE}},
  \bibinfo{pages}{200--210}.
\newblock


\bibitem[\protect\citeauthoryear{Feichtenhofer, Fan, Malik, and
  He}{Feichtenhofer et~al\mbox{.}}{2019}]%
        {DBLP:conf/iccv/Feichtenhofer0M19}
\bibfield{author}{\bibinfo{person}{Christoph Feichtenhofer},
  \bibinfo{person}{Haoqi Fan}, \bibinfo{person}{Jitendra Malik}, {and}
  \bibinfo{person}{Kaiming He}.} \bibinfo{year}{2019}\natexlab{}.
\newblock \showarticletitle{SlowFast Networks for Video Recognition}. In
  \bibinfo{booktitle}{\emph{{IEEE/CVF} International Conference on Computer
  Vision, {ICCV}}}. \bibinfo{publisher}{{IEEE}}, \bibinfo{pages}{6201--6210}.
\newblock


\bibitem[\protect\citeauthoryear{Feichtenhofer, Pinz, and Wildes}{Feichtenhofer
  et~al\mbox{.}}{2016a}]%
        {DBLP:conf/nips/FeichtenhoferPW16}
\bibfield{author}{\bibinfo{person}{Christoph Feichtenhofer},
  \bibinfo{person}{Axel Pinz}, {and} \bibinfo{person}{Richard~P. Wildes}.}
  \bibinfo{year}{2016}\natexlab{a}.
\newblock \showarticletitle{Spatiotemporal Residual Networks for Video Action
  Recognition}. In \bibinfo{booktitle}{\emph{Advances in Neural Information
  Processing Systems 29: Annual Conference on Neural Information Processing
  Systems}}. \bibinfo{pages}{3468--3476}.
\newblock


\bibitem[\protect\citeauthoryear{Feichtenhofer, Pinz, and Wildes}{Feichtenhofer
  et~al\mbox{.}}{2017}]%
        {DBLP:conf/cvpr/FeichtenhoferPW17a}
\bibfield{author}{\bibinfo{person}{Christoph Feichtenhofer},
  \bibinfo{person}{Axel Pinz}, {and} \bibinfo{person}{Richard~P. Wildes}.}
  \bibinfo{year}{2017}\natexlab{}.
\newblock \showarticletitle{Spatiotemporal Multiplier Networks for Video Action
  Recognition}. In \bibinfo{booktitle}{\emph{{IEEE} Conference on Computer
  Vision and Pattern Recognition, {CVPR}}}. \bibinfo{publisher}{{IEEE} Computer
  Society}, \bibinfo{pages}{7445--7454}.
\newblock


\bibitem[\protect\citeauthoryear{Feichtenhofer, Pinz, and
  Zisserman}{Feichtenhofer et~al\mbox{.}}{2016b}]%
        {DBLP:conf/cvpr/FeichtenhoferPZ16}
\bibfield{author}{\bibinfo{person}{Christoph Feichtenhofer},
  \bibinfo{person}{Axel Pinz}, {and} \bibinfo{person}{Andrew Zisserman}.}
  \bibinfo{year}{2016}\natexlab{b}.
\newblock \showarticletitle{Convolutional Two-Stream Network Fusion for Video
  Action Recognition}. In \bibinfo{booktitle}{\emph{{IEEE} Conference on
  Computer Vision and Pattern Recognition, {CVPR}}}. \bibinfo{publisher}{{IEEE}
  Computer Society}, \bibinfo{pages}{1933--1941}.
\newblock


\bibitem[\protect\citeauthoryear{Goyal, Ebrahimi~Kahou, Michalski, Materzynska,
  Westphal, Kim, Haenel, Fruend, Yianilos, Mueller-Freitag,
  et~al\mbox{.}}{Goyal et~al\mbox{.}}{2017}]%
        {goyal2017something}
\bibfield{author}{\bibinfo{person}{Raghav Goyal}, \bibinfo{person}{Samira
  Ebrahimi~Kahou}, \bibinfo{person}{Vincent Michalski}, \bibinfo{person}{Joanna
  Materzynska}, \bibinfo{person}{Susanne Westphal}, \bibinfo{person}{Heuna
  Kim}, \bibinfo{person}{Valentin Haenel}, \bibinfo{person}{Ingo Fruend},
  \bibinfo{person}{Peter Yianilos}, \bibinfo{person}{Moritz Mueller-Freitag},
  {et~al\mbox{.}}} \bibinfo{year}{2017}\natexlab{}.
\newblock \showarticletitle{The" something something" video database for
  learning and evaluating visual common sense}. In
  \bibinfo{booktitle}{\emph{Proceedings of the IEEE International Conference on
  Computer Vision}}. \bibinfo{pages}{5842--5850}.
\newblock


\bibitem[\protect\citeauthoryear{Hara, Kataoka, and Satoh}{Hara
  et~al\mbox{.}}{2018}]%
        {DBLP:conf/cvpr/HaraKS18}
\bibfield{author}{\bibinfo{person}{Kensho Hara}, \bibinfo{person}{Hirokatsu
  Kataoka}, {and} \bibinfo{person}{Yutaka Satoh}.}
  \bibinfo{year}{2018}\natexlab{}.
\newblock \showarticletitle{Can Spatiotemporal 3D CNNs Retrace the History of
  2D CNNs and ImageNet?}. In \bibinfo{booktitle}{\emph{{IEEE} Conference on
  Computer Vision and Pattern Recognition, {CVPR}}}. \bibinfo{publisher}{{IEEE}
  Computer Society}, \bibinfo{pages}{6546--6555}.
\newblock


\bibitem[\protect\citeauthoryear{Hinton, Vinyals, and Dean}{Hinton
  et~al\mbox{.}}{2015}]%
        {hinton2015distilling}
\bibfield{author}{\bibinfo{person}{Geoffrey Hinton}, \bibinfo{person}{Oriol
  Vinyals}, {and} \bibinfo{person}{Jeff Dean}.}
  \bibinfo{year}{2015}\natexlab{}.
\newblock \showarticletitle{Distilling the knowledge in a neural network}.
\newblock \bibinfo{journal}{\emph{arXiv preprint arXiv:1503.02531}}
  (\bibinfo{year}{2015}).
\newblock


\bibitem[\protect\citeauthoryear{Hou, Pan, Loy, Wang, and Lin}{Hou
  et~al\mbox{.}}{2019}]%
        {hou2019learning}
\bibfield{author}{\bibinfo{person}{Saihui Hou}, \bibinfo{person}{Xinyu Pan},
  \bibinfo{person}{Chen~Change Loy}, \bibinfo{person}{Zilei Wang}, {and}
  \bibinfo{person}{Dahua Lin}.} \bibinfo{year}{2019}\natexlab{}.
\newblock \showarticletitle{Learning a unified classifier incrementally via
  rebalancing}. In \bibinfo{booktitle}{\emph{Proceedings of the IEEE/CVF
  Conference on Computer Vision and Pattern Recognition}}.
  \bibinfo{pages}{831--839}.
\newblock


\bibitem[\protect\citeauthoryear{Iscen, Zhang, Lazebnik, and Schmid}{Iscen
  et~al\mbox{.}}{2020}]%
        {iscen2020memory}
\bibfield{author}{\bibinfo{person}{Ahmet Iscen}, \bibinfo{person}{Jeffrey
  Zhang}, \bibinfo{person}{Svetlana Lazebnik}, {and} \bibinfo{person}{Cordelia
  Schmid}.} \bibinfo{year}{2020}\natexlab{}.
\newblock \showarticletitle{Memory-efficient incremental learning through
  feature adaptation}. In \bibinfo{booktitle}{\emph{European Conference on
  Computer Vision}}. Springer, \bibinfo{pages}{699--715}.
\newblock


\bibitem[\protect\citeauthoryear{Ji, Xu, Yang, and Yu}{Ji
  et~al\mbox{.}}{2013}]%
        {DBLP:journals/pami/JiXYY13}
\bibfield{author}{\bibinfo{person}{Shuiwang Ji}, \bibinfo{person}{Wei Xu},
  \bibinfo{person}{Ming Yang}, {and} \bibinfo{person}{Kai Yu}.}
  \bibinfo{year}{2013}\natexlab{}.
\newblock \showarticletitle{3D Convolutional Neural Networks for Human Action
  Recognition}.
\newblock \bibinfo{journal}{\emph{{IEEE} Trans. Pattern Anal. Mach. Intell.}}
  \bibinfo{volume}{35}, \bibinfo{number}{1} (\bibinfo{year}{2013}),
  \bibinfo{pages}{221--231}.
\newblock


\bibitem[\protect\citeauthoryear{Ji, Li, Wu, Pan, and Zhuang}{Ji
  et~al\mbox{.}}{2019}]%
        {ji2019human}
\bibfield{author}{\bibinfo{person}{Wei Ji}, \bibinfo{person}{Xi Li},
  \bibinfo{person}{Fei Wu}, \bibinfo{person}{Zhijie Pan}, {and}
  \bibinfo{person}{Yueting Zhuang}.} \bibinfo{year}{2019}\natexlab{}.
\newblock \showarticletitle{Human-centric clothing segmentation via deformable
  semantic locality-preserving network}.
\newblock \bibinfo{journal}{\emph{IEEE Transactions on Circuits and Systems for
  Video Technology}} \bibinfo{volume}{30}, \bibinfo{number}{12}
  (\bibinfo{year}{2019}), \bibinfo{pages}{4837--4848}.
\newblock


\bibitem[\protect\citeauthoryear{Jiang, Wang, Gan, Wu, and Yan}{Jiang
  et~al\mbox{.}}{2019}]%
        {DBLP:conf/iccv/JiangWGWY19}
\bibfield{author}{\bibinfo{person}{Boyuan Jiang}, \bibinfo{person}{Mengmeng
  Wang}, \bibinfo{person}{Weihao Gan}, \bibinfo{person}{Wei Wu}, {and}
  \bibinfo{person}{Junjie Yan}.} \bibinfo{year}{2019}\natexlab{}.
\newblock \showarticletitle{{STM:} SpatioTemporal and Motion Encoding for
  Action Recognition}. In \bibinfo{booktitle}{\emph{{IEEE/CVF} International
  Conference on Computer Vision, {ICCV}}}. \bibinfo{publisher}{{IEEE}},
  \bibinfo{pages}{2000--2009}.
\newblock


\bibitem[\protect\citeauthoryear{Kamra, Gupta, and Liu}{Kamra
  et~al\mbox{.}}{2017}]%
        {kamra2017deep}
\bibfield{author}{\bibinfo{person}{Nitin Kamra}, \bibinfo{person}{Umang Gupta},
  {and} \bibinfo{person}{Yan Liu}.} \bibinfo{year}{2017}\natexlab{}.
\newblock \showarticletitle{Deep generative dual memory network for continual
  learning}.
\newblock \bibinfo{journal}{\emph{arXiv preprint arXiv:1710.10368}}
  (\bibinfo{year}{2017}).
\newblock


\bibitem[\protect\citeauthoryear{Kay, Carreira, Simonyan, Zhang, Hillier,
  Vijayanarasimhan, Viola, Green, Back, Natsev, et~al\mbox{.}}{Kay
  et~al\mbox{.}}{2017}]%
        {kay2017kinetics}
\bibfield{author}{\bibinfo{person}{Will Kay}, \bibinfo{person}{Joao Carreira},
  \bibinfo{person}{Karen Simonyan}, \bibinfo{person}{Brian Zhang},
  \bibinfo{person}{Chloe Hillier}, \bibinfo{person}{Sudheendra
  Vijayanarasimhan}, \bibinfo{person}{Fabio Viola}, \bibinfo{person}{Tim
  Green}, \bibinfo{person}{Trevor Back}, \bibinfo{person}{Paul Natsev},
  {et~al\mbox{.}}} \bibinfo{year}{2017}\natexlab{}.
\newblock \showarticletitle{The kinetics human action video dataset}.
\newblock \bibinfo{journal}{\emph{arXiv preprint arXiv:1705.06950}}
  (\bibinfo{year}{2017}).
\newblock


\bibitem[\protect\citeauthoryear{Kwon, Kim, Kwak, and Cho}{Kwon
  et~al\mbox{.}}{2020}]%
        {kwon2020motionsqueeze}
\bibfield{author}{\bibinfo{person}{Heeseung Kwon}, \bibinfo{person}{Manjin
  Kim}, \bibinfo{person}{Suha Kwak}, {and} \bibinfo{person}{Minsu Cho}.}
  \bibinfo{year}{2020}\natexlab{}.
\newblock \showarticletitle{Motionsqueeze: Neural motion feature learning for
  video understanding}. In \bibinfo{booktitle}{\emph{European Conference on
  Computer Vision}}. Springer, \bibinfo{pages}{345--362}.
\newblock


\bibitem[\protect\citeauthoryear{Lan, Lin, Li, Hauptmann, and Raj}{Lan
  et~al\mbox{.}}{2015}]%
        {DBLP:conf/cvpr/LanLLHR15}
\bibfield{author}{\bibinfo{person}{Zhen{-}Zhong Lan}, \bibinfo{person}{Ming
  Lin}, \bibinfo{person}{Xuanchong Li}, \bibinfo{person}{Alexander~G.
  Hauptmann}, {and} \bibinfo{person}{Bhiksha Raj}.}
  \bibinfo{year}{2015}\natexlab{}.
\newblock \showarticletitle{Beyond Gaussian Pyramid: Multi-skip Feature
  Stacking for action recognition}. In \bibinfo{booktitle}{\emph{{IEEE}
  Conference on Computer Vision and Pattern Recognition, {CVPR}}}.
  \bibinfo{publisher}{{IEEE} Computer Society}, \bibinfo{pages}{204--212}.
\newblock


\bibitem[\protect\citeauthoryear{Li, Zhao, Ji, Wu, Wu, Yang, Tao, and Reid}{Li
  et~al\mbox{.}}{2018}]%
        {li2018multi}
\bibfield{author}{\bibinfo{person}{Xi Li}, \bibinfo{person}{Liming Zhao},
  \bibinfo{person}{Wei Ji}, \bibinfo{person}{Yiming Wu}, \bibinfo{person}{Fei
  Wu}, \bibinfo{person}{Ming-Hsuan Yang}, \bibinfo{person}{Dacheng Tao}, {and}
  \bibinfo{person}{Ian Reid}.} \bibinfo{year}{2018}\natexlab{}.
\newblock \showarticletitle{Multi-task structure-aware context modeling for
  robust keypoint-based object tracking}.
\newblock \bibinfo{journal}{\emph{IEEE transactions on pattern analysis and
  machine intelligence}} \bibinfo{volume}{41}, \bibinfo{number}{4}
  (\bibinfo{year}{2018}), \bibinfo{pages}{915--927}.
\newblock


\bibitem[\protect\citeauthoryear{Li, Ji, Shi, Zhang, Kang, and Wang}{Li
  et~al\mbox{.}}{2020}]%
        {DBLP:conf/cvpr/LiJSZKW20}
\bibfield{author}{\bibinfo{person}{Yan Li}, \bibinfo{person}{Bin Ji},
  \bibinfo{person}{Xintian Shi}, \bibinfo{person}{Jianguo Zhang},
  \bibinfo{person}{Bin Kang}, {and} \bibinfo{person}{Limin Wang}.}
  \bibinfo{year}{2020}\natexlab{}.
\newblock \showarticletitle{{TEA:} Temporal Excitation and Aggregation for
  Action Recognition}. In \bibinfo{booktitle}{\emph{{IEEE/CVF} Conference on
  Computer Vision and Pattern Recognition, {CVPR}}}.
  \bibinfo{publisher}{{IEEE}}, \bibinfo{pages}{906--915}.
\newblock


\bibitem[\protect\citeauthoryear{Li and Hoiem}{Li and Hoiem}{2017}]%
        {li2017learning}
\bibfield{author}{\bibinfo{person}{Zhizhong Li} {and} \bibinfo{person}{Derek
  Hoiem}.} \bibinfo{year}{2017}\natexlab{}.
\newblock \showarticletitle{Learning without forgetting}.
\newblock \bibinfo{journal}{\emph{IEEE transactions on pattern analysis and
  machine intelligence}} \bibinfo{volume}{40}, \bibinfo{number}{12}
  (\bibinfo{year}{2017}), \bibinfo{pages}{2935--2947}.
\newblock


\bibitem[\protect\citeauthoryear{Lin, Gan, and Han}{Lin et~al\mbox{.}}{2019}]%
        {lin2019tsm}
\bibfield{author}{\bibinfo{person}{Ji Lin}, \bibinfo{person}{Chuang Gan}, {and}
  \bibinfo{person}{Song Han}.} \bibinfo{year}{2019}\natexlab{}.
\newblock \showarticletitle{Tsm: Temporal shift module for efficient video
  understanding}. In \bibinfo{booktitle}{\emph{Proceedings of the IEEE/CVF
  International Conference on Computer Vision}}. \bibinfo{pages}{7083--7093}.
\newblock


\bibitem[\protect\citeauthoryear{Liu, Su, Liu, Schiele, and Sun}{Liu
  et~al\mbox{.}}{2020b}]%
        {liu2020mnemonics}
\bibfield{author}{\bibinfo{person}{Yaoyao Liu}, \bibinfo{person}{Yuting Su},
  \bibinfo{person}{An-An Liu}, \bibinfo{person}{Bernt Schiele}, {and}
  \bibinfo{person}{Qianru Sun}.} \bibinfo{year}{2020}\natexlab{b}.
\newblock \showarticletitle{Mnemonics training: Multi-class incremental
  learning without forgetting}. In \bibinfo{booktitle}{\emph{Proceedings of the
  IEEE/CVF Conference on Computer Vision and Pattern Recognition}}.
  \bibinfo{pages}{12245--12254}.
\newblock


\bibitem[\protect\citeauthoryear{Liu, Luo, Wang, Wang, Tai, Wang, Li, Huang,
  and Lu}{Liu et~al\mbox{.}}{2020a}]%
        {DBLP:conf/aaai/LiuLWWTWLHL20}
\bibfield{author}{\bibinfo{person}{Zhaoyang Liu}, \bibinfo{person}{Donghao
  Luo}, \bibinfo{person}{Yabiao Wang}, \bibinfo{person}{Limin Wang},
  \bibinfo{person}{Ying Tai}, \bibinfo{person}{Chengjie Wang},
  \bibinfo{person}{Jilin Li}, \bibinfo{person}{Feiyue Huang}, {and}
  \bibinfo{person}{Tong Lu}.} \bibinfo{year}{2020}\natexlab{a}.
\newblock \showarticletitle{TEINet: Towards an Efficient Architecture for Video
  Recognition}. In \bibinfo{booktitle}{\emph{{AAAI} Conference on Artificial
  Intelligence}}. \bibinfo{publisher}{{AAAI} Press},
  \bibinfo{pages}{11669--11676}.
\newblock


\bibitem[\protect\citeauthoryear{McCloskey and Cohen}{McCloskey and
  Cohen}{1989}]%
        {mccloskey1989catastrophic}
\bibfield{author}{\bibinfo{person}{Michael McCloskey} {and}
  \bibinfo{person}{Neal~J Cohen}.} \bibinfo{year}{1989}\natexlab{}.
\newblock \showarticletitle{Catastrophic interference in connectionist
  networks: The sequential learning problem}.
\newblock In \bibinfo{booktitle}{\emph{Psychology of learning and motivation}}.
  Vol.~\bibinfo{volume}{24}. \bibinfo{publisher}{Elsevier},
  \bibinfo{pages}{109--165}.
\newblock


\bibitem[\protect\citeauthoryear{Paszke, Gross, Chintala, Chanan, Yang, DeVito,
  Lin, Desmaison, Antiga, and Lerer}{Paszke et~al\mbox{.}}{2017}]%
        {paszke2017automatic}
\bibfield{author}{\bibinfo{person}{Adam Paszke}, \bibinfo{person}{Sam Gross},
  \bibinfo{person}{Soumith Chintala}, \bibinfo{person}{Gregory Chanan},
  \bibinfo{person}{Edward Yang}, \bibinfo{person}{Zachary DeVito},
  \bibinfo{person}{Zeming Lin}, \bibinfo{person}{Alban Desmaison},
  \bibinfo{person}{Luca Antiga}, {and} \bibinfo{person}{Adam Lerer}.}
  \bibinfo{year}{2017}\natexlab{}.
\newblock \showarticletitle{Automatic differentiation in pytorch}.
\newblock  (\bibinfo{year}{2017}).
\newblock


\bibitem[\protect\citeauthoryear{Peng, Zou, Qiao, and Peng}{Peng
  et~al\mbox{.}}{2014}]%
        {DBLP:conf/eccv/PengZQP14}
\bibfield{author}{\bibinfo{person}{Xiaojiang Peng}, \bibinfo{person}{Changqing
  Zou}, \bibinfo{person}{Yu Qiao}, {and} \bibinfo{person}{Qiang Peng}.}
  \bibinfo{year}{2014}\natexlab{}.
\newblock \showarticletitle{Action Recognition with Stacked Fisher Vectors}. In
  \bibinfo{booktitle}{\emph{European Conference of Computer Vision , {ECCV}}}
  \emph{(\bibinfo{series}{Lecture Notes in Computer Science},
  Vol.~\bibinfo{volume}{8693})}. \bibinfo{publisher}{Springer},
  \bibinfo{pages}{581--595}.
\newblock


\bibitem[\protect\citeauthoryear{Qin, Zhao, Lin, Zeng, Xu, and Li}{Qin
  et~al\mbox{.}}{2021}]%
        {qin2021pcmnet}
\bibfield{author}{\bibinfo{person}{Xin Qin}, \bibinfo{person}{Hanbin Zhao},
  \bibinfo{person}{Guangchen Lin}, \bibinfo{person}{Hao Zeng},
  \bibinfo{person}{Songcen Xu}, {and} \bibinfo{person}{Xi Li}.}
  \bibinfo{year}{2021}\natexlab{}.
\newblock \showarticletitle{PcmNet: Position-Sensitive Context Modeling Network
  for Temporal Action Localization}.
\newblock \bibinfo{journal}{\emph{arXiv preprint arXiv:2103.05270}}
  (\bibinfo{year}{2021}).
\newblock


\bibitem[\protect\citeauthoryear{Qiu, Yao, and Mei}{Qiu et~al\mbox{.}}{2017}]%
        {DBLP:conf/iccv/QiuYM17}
\bibfield{author}{\bibinfo{person}{Zhaofan Qiu}, \bibinfo{person}{Ting Yao},
  {and} \bibinfo{person}{Tao Mei}.} \bibinfo{year}{2017}\natexlab{}.
\newblock \showarticletitle{Learning Spatio-Temporal Representation with
  Pseudo-3D Residual Networks}. In \bibinfo{booktitle}{\emph{{IEEE}
  International Conference on Computer Vision, {ICCV}}}.
  \bibinfo{publisher}{{IEEE} Computer Society}, \bibinfo{pages}{5534--5542}.
\newblock


\bibitem[\protect\citeauthoryear{Rebuffi, Kolesnikov, Sperl, and
  Lampert}{Rebuffi et~al\mbox{.}}{2017}]%
        {rebuffi2017icarl}
\bibfield{author}{\bibinfo{person}{Sylvestre-Alvise Rebuffi},
  \bibinfo{person}{Alexander Kolesnikov}, \bibinfo{person}{Georg Sperl}, {and}
  \bibinfo{person}{Christoph~H Lampert}.} \bibinfo{year}{2017}\natexlab{}.
\newblock \showarticletitle{icarl: Incremental classifier and representation
  learning}. In \bibinfo{booktitle}{\emph{Proceedings of the IEEE conference on
  Computer Vision and Pattern Recognition}}. \bibinfo{pages}{2001--2010}.
\newblock


\bibitem[\protect\citeauthoryear{Shao, Qian, and Liu}{Shao
  et~al\mbox{.}}{2020}]%
        {DBLP:conf/aaai/ShaoQL20}
\bibfield{author}{\bibinfo{person}{Hao Shao}, \bibinfo{person}{Shengju Qian},
  {and} \bibinfo{person}{Yu Liu}.} \bibinfo{year}{2020}\natexlab{}.
\newblock \showarticletitle{Temporal Interlacing Network}. In
  \bibinfo{booktitle}{\emph{{AAAI} Conference on Artificial Intelligence}}.
  \bibinfo{publisher}{{AAAI} Press}, \bibinfo{pages}{11966--11973}.
\newblock


\bibitem[\protect\citeauthoryear{Shin, Lee, Kim, and Kim}{Shin
  et~al\mbox{.}}{2017}]%
        {shin2017continual}
\bibfield{author}{\bibinfo{person}{Hanul Shin}, \bibinfo{person}{Jung~Kwon
  Lee}, \bibinfo{person}{Jaehong Kim}, {and} \bibinfo{person}{Jiwon Kim}.}
  \bibinfo{year}{2017}\natexlab{}.
\newblock \showarticletitle{Continual learning with deep generative replay}.
\newblock \bibinfo{journal}{\emph{arXiv preprint arXiv:1705.08690}}
  (\bibinfo{year}{2017}).
\newblock


\bibitem[\protect\citeauthoryear{Simonyan and Zisserman}{Simonyan and
  Zisserman}{2014}]%
        {DBLP:conf/nips/SimonyanZ14}
\bibfield{author}{\bibinfo{person}{Karen Simonyan} {and}
  \bibinfo{person}{Andrew Zisserman}.} \bibinfo{year}{2014}\natexlab{}.
\newblock \showarticletitle{Two-Stream Convolutional Networks for Action
  Recognition in Videos}. In \bibinfo{booktitle}{\emph{Advances in Neural
  Information Processing Systems 27: Annual Conference on Neural Information
  Processing Systems}}. \bibinfo{pages}{568--576}.
\newblock


\bibitem[\protect\citeauthoryear{Tao, Chang, Hong, Wei, and Gong}{Tao
  et~al\mbox{.}}{2020}]%
        {tao2020topology}
\bibfield{author}{\bibinfo{person}{Xiaoyu Tao}, \bibinfo{person}{Xinyuan
  Chang}, \bibinfo{person}{Xiaopeng Hong}, \bibinfo{person}{Xing Wei}, {and}
  \bibinfo{person}{Yihong Gong}.} \bibinfo{year}{2020}\natexlab{}.
\newblock \showarticletitle{Topology-preserving class-incremental learning}. In
  \bibinfo{booktitle}{\emph{European Conference on Computer Vision}}. Springer,
  \bibinfo{pages}{254--270}.
\newblock


\bibitem[\protect\citeauthoryear{Tran, Bourdev, Fergus, Torresani, and
  Paluri}{Tran et~al\mbox{.}}{2015}]%
        {DBLP:conf/iccv/TranBFTP15}
\bibfield{author}{\bibinfo{person}{Du Tran}, \bibinfo{person}{Lubomir~D.
  Bourdev}, \bibinfo{person}{Rob Fergus}, \bibinfo{person}{Lorenzo Torresani},
  {and} \bibinfo{person}{Manohar Paluri}.} \bibinfo{year}{2015}\natexlab{}.
\newblock \showarticletitle{Learning Spatiotemporal Features with 3D
  Convolutional Networks}. In \bibinfo{booktitle}{\emph{{IEEE} International
  Conference on Computer Vision, {ICCV} 2015}}. \bibinfo{publisher}{{IEEE}
  Computer Society}, \bibinfo{pages}{4489--4497}.
\newblock


\bibitem[\protect\citeauthoryear{Tran, Wang, Torresani, and Feiszli}{Tran
  et~al\mbox{.}}{2019}]%
        {tran2019video}
\bibfield{author}{\bibinfo{person}{Du Tran}, \bibinfo{person}{Heng Wang},
  \bibinfo{person}{Lorenzo Torresani}, {and} \bibinfo{person}{Matt Feiszli}.}
  \bibinfo{year}{2019}\natexlab{}.
\newblock \showarticletitle{Video classification with channel-separated
  convolutional networks}. In \bibinfo{booktitle}{\emph{Proceedings of the
  IEEE/CVF International Conference on Computer Vision}}.
  \bibinfo{pages}{5552--5561}.
\newblock


\bibitem[\protect\citeauthoryear{Tran, Wang, Torresani, Ray, LeCun, and
  Paluri}{Tran et~al\mbox{.}}{2018}]%
        {DBLP:conf/cvpr/TranWTRLP18}
\bibfield{author}{\bibinfo{person}{Du Tran}, \bibinfo{person}{Heng Wang},
  \bibinfo{person}{Lorenzo Torresani}, \bibinfo{person}{Jamie Ray},
  \bibinfo{person}{Yann LeCun}, {and} \bibinfo{person}{Manohar Paluri}.}
  \bibinfo{year}{2018}\natexlab{}.
\newblock \showarticletitle{A Closer Look at Spatiotemporal Convolutions for
  Action Recognition}. In \bibinfo{booktitle}{\emph{{IEEE} Conference on
  Computer Vision and Pattern Recognition, {CVPR}}}. \bibinfo{publisher}{{IEEE}
  Computer Society}, \bibinfo{pages}{6450--6459}.
\newblock


\bibitem[\protect\citeauthoryear{Wang, Kl{\"{a}}ser, Schmid, and Liu}{Wang
  et~al\mbox{.}}{2011}]%
        {DBLP:conf/cvpr/WangKSL11}
\bibfield{author}{\bibinfo{person}{Heng Wang}, \bibinfo{person}{Alexander
  Kl{\"{a}}ser}, \bibinfo{person}{Cordelia Schmid}, {and}
  \bibinfo{person}{Cheng{-}Lin Liu}.} \bibinfo{year}{2011}\natexlab{}.
\newblock \showarticletitle{Action recognition by dense trajectories}. In
  \bibinfo{booktitle}{\emph{T{IEEE} Conference on Computer Vision and Pattern
  Recognition, {CVPR}}}. \bibinfo{publisher}{{IEEE} Computer Society},
  \bibinfo{pages}{3169--3176}.
\newblock


\bibitem[\protect\citeauthoryear{Wang and Schmid}{Wang and Schmid}{2013}]%
        {wang2013action}
\bibfield{author}{\bibinfo{person}{Heng Wang} {and} \bibinfo{person}{Cordelia
  Schmid}.} \bibinfo{year}{2013}\natexlab{}.
\newblock \showarticletitle{Action Recognition with Improved Trajectories}. In
  \bibinfo{booktitle}{\emph{{IEEE} International Conference on Computer Vision,
  {ICCV}}}. \bibinfo{publisher}{{IEEE} Computer Society},
  \bibinfo{pages}{3551--3558}.
\newblock


\bibitem[\protect\citeauthoryear{Wang, Li, Li, and Gool}{Wang
  et~al\mbox{.}}{2018b}]%
        {DBLP:conf/cvpr/WangL0G18}
\bibfield{author}{\bibinfo{person}{Limin Wang}, \bibinfo{person}{Wei Li},
  \bibinfo{person}{Wen Li}, {and} \bibinfo{person}{Luc~Van Gool}.}
  \bibinfo{year}{2018}\natexlab{b}.
\newblock \showarticletitle{Appearance-and-Relation Networks for Video
  Classification}. In \bibinfo{booktitle}{\emph{{IEEE} Conference on Computer
  Vision and Pattern Recognition, {CVPR}}}. \bibinfo{publisher}{{IEEE} Computer
  Society}, \bibinfo{pages}{1430--1439}.
\newblock


\bibitem[\protect\citeauthoryear{Wang, Qiao, and Tang}{Wang
  et~al\mbox{.}}{2015}]%
        {wang2015action}
\bibfield{author}{\bibinfo{person}{Limin Wang}, \bibinfo{person}{Yu Qiao},
  {and} \bibinfo{person}{Xiaoou Tang}.} \bibinfo{year}{2015}\natexlab{}.
\newblock \showarticletitle{Action recognition with trajectory-pooled
  deep-convolutional descriptors}. In \bibinfo{booktitle}{\emph{Proceedings of
  the IEEE conference on computer vision and pattern recognition}}.
  \bibinfo{pages}{4305--4314}.
\newblock


\bibitem[\protect\citeauthoryear{Wang, Xiong, Wang, Qiao, Lin, Tang, and
  Van~Gool}{Wang et~al\mbox{.}}{2016}]%
        {wang2016temporal}
\bibfield{author}{\bibinfo{person}{Limin Wang}, \bibinfo{person}{Yuanjun
  Xiong}, \bibinfo{person}{Zhe Wang}, \bibinfo{person}{Yu Qiao},
  \bibinfo{person}{Dahua Lin}, \bibinfo{person}{Xiaoou Tang}, {and}
  \bibinfo{person}{Luc Van~Gool}.} \bibinfo{year}{2016}\natexlab{}.
\newblock \showarticletitle{Temporal segment networks: Towards good practices
  for deep action recognition}. In \bibinfo{booktitle}{\emph{European
  conference on computer vision}}. Springer, \bibinfo{pages}{20--36}.
\newblock


\bibitem[\protect\citeauthoryear{Wang, Girshick, Gupta, and He}{Wang
  et~al\mbox{.}}{2018a}]%
        {DBLP:conf/cvpr/0004GGH18}
\bibfield{author}{\bibinfo{person}{Xiaolong Wang}, \bibinfo{person}{Ross~B.
  Girshick}, \bibinfo{person}{Abhinav Gupta}, {and} \bibinfo{person}{Kaiming
  He}.} \bibinfo{year}{2018}\natexlab{a}.
\newblock \showarticletitle{Non-Local Neural Networks}. In
  \bibinfo{booktitle}{\emph{{IEEE} Conference on Computer Vision and Pattern
  Recognition, {CVPR}}}. \bibinfo{publisher}{{IEEE} Computer Society},
  \bibinfo{pages}{7794--7803}.
\newblock


\bibitem[\protect\citeauthoryear{Wang, Long, Wang, and Yu}{Wang
  et~al\mbox{.}}{2017}]%
        {DBLP:conf/cvpr/WangLWY17}
\bibfield{author}{\bibinfo{person}{Yunbo Wang}, \bibinfo{person}{Mingsheng
  Long}, \bibinfo{person}{Jianmin Wang}, {and} \bibinfo{person}{Philip~S. Yu}.}
  \bibinfo{year}{2017}\natexlab{}.
\newblock \showarticletitle{Spatiotemporal Pyramid Network for Video Action
  Recognition}. In \bibinfo{booktitle}{\emph{{IEEE} Conference on Computer
  Vision and Pattern Recognition, {CVPR}}}. \bibinfo{publisher}{{IEEE} Computer
  Society}, \bibinfo{pages}{2097--2106}.
\newblock


\bibitem[\protect\citeauthoryear{Welling}{Welling}{2009}]%
        {welling2009herding}
\bibfield{author}{\bibinfo{person}{Max Welling}.}
  \bibinfo{year}{2009}\natexlab{}.
\newblock \showarticletitle{Herding dynamical weights to learn}. In
  \bibinfo{booktitle}{\emph{Proceedings of the 26th Annual International
  Conference on Machine Learning}}. \bibinfo{pages}{1121--1128}.
\newblock


\bibitem[\protect\citeauthoryear{Wu, Bourahla, Li, Wu, Tian, and Zhou}{Wu
  et~al\mbox{.}}{2020}]%
        {wu2020adaptive}
\bibfield{author}{\bibinfo{person}{Yiming Wu}, \bibinfo{person}{Omar El~Farouk
  Bourahla}, \bibinfo{person}{Xi Li}, \bibinfo{person}{Fei Wu},
  \bibinfo{person}{Qi Tian}, {and} \bibinfo{person}{Xue Zhou}.}
  \bibinfo{year}{2020}\natexlab{}.
\newblock \showarticletitle{Adaptive graph representation learning for video
  person re-identification}.
\newblock \bibinfo{journal}{\emph{IEEE Transactions on Image Processing}}
  \bibinfo{volume}{29} (\bibinfo{year}{2020}), \bibinfo{pages}{8821--8830}.
\newblock


\bibitem[\protect\citeauthoryear{Wu, Chen, Wang, Ye, Liu, Guo, and Fu}{Wu
  et~al\mbox{.}}{2019}]%
        {wu2019large}
\bibfield{author}{\bibinfo{person}{Yue Wu}, \bibinfo{person}{Yinpeng Chen},
  \bibinfo{person}{Lijuan Wang}, \bibinfo{person}{Yuancheng Ye},
  \bibinfo{person}{Zicheng Liu}, \bibinfo{person}{Yandong Guo}, {and}
  \bibinfo{person}{Yun Fu}.} \bibinfo{year}{2019}\natexlab{}.
\newblock \showarticletitle{Large scale incremental learning}. In
  \bibinfo{booktitle}{\emph{Proceedings of the IEEE/CVF Conference on Computer
  Vision and Pattern Recognition}}. \bibinfo{pages}{374--382}.
\newblock


\bibitem[\protect\citeauthoryear{Wu, Ji, Li, Wang, Yin, and Wu}{Wu
  et~al\mbox{.}}{2018}]%
        {wu2018context}
\bibfield{author}{\bibinfo{person}{Yiming Wu}, \bibinfo{person}{Wei Ji},
  \bibinfo{person}{Xi Li}, \bibinfo{person}{Gang Wang},
  \bibinfo{person}{Jianwei Yin}, {and} \bibinfo{person}{Fei Wu}.}
  \bibinfo{year}{2018}\natexlab{}.
\newblock \showarticletitle{Context-aware deep spatiotemporal network for hand
  pose estimation from depth images}.
\newblock \bibinfo{journal}{\emph{IEEE transactions on cybernetics}}
  \bibinfo{volume}{50}, \bibinfo{number}{2} (\bibinfo{year}{2018}),
  \bibinfo{pages}{787--797}.
\newblock


\bibitem[\protect\citeauthoryear{Wu, Jiang, Wang, Ye, and Xue}{Wu
  et~al\mbox{.}}{2016}]%
        {DBLP:conf/mm/WuJWYX16}
\bibfield{author}{\bibinfo{person}{Zuxuan Wu}, \bibinfo{person}{Yu{-}Gang
  Jiang}, \bibinfo{person}{Xi Wang}, \bibinfo{person}{Hao Ye}, {and}
  \bibinfo{person}{Xiangyang Xue}.} \bibinfo{year}{2016}\natexlab{}.
\newblock \showarticletitle{Multi-Stream Multi-Class Fusion of Deep Networks
  for Video Classification}. In \bibinfo{booktitle}{\emph{{ACM} Conference on
  Multimedia Conference, {MM}}}. \bibinfo{publisher}{{ACM}},
  \bibinfo{pages}{791--800}.
\newblock


\bibitem[\protect\citeauthoryear{Xie, Sun, Huang, Tu, and Murphy}{Xie
  et~al\mbox{.}}{2018a}]%
        {xie2018rethinking}
\bibfield{author}{\bibinfo{person}{Saining Xie}, \bibinfo{person}{Chen Sun},
  \bibinfo{person}{Jonathan Huang}, \bibinfo{person}{Zhuowen Tu}, {and}
  \bibinfo{person}{Kevin Murphy}.} \bibinfo{year}{2018}\natexlab{a}.
\newblock \showarticletitle{Rethinking spatiotemporal feature learning:
  Speed-accuracy trade-offs in video classification}. In
  \bibinfo{booktitle}{\emph{Proceedings of the European Conference on Computer
  Vision (ECCV)}}. \bibinfo{pages}{305--321}.
\newblock


\bibitem[\protect\citeauthoryear{Xie, Sun, Huang, Tu, and Murphy}{Xie
  et~al\mbox{.}}{2018b}]%
        {DBLP:conf/eccv/XieSHTM18}
\bibfield{author}{\bibinfo{person}{Saining Xie}, \bibinfo{person}{Chen Sun},
  \bibinfo{person}{Jonathan Huang}, \bibinfo{person}{Zhuowen Tu}, {and}
  \bibinfo{person}{Kevin Murphy}.} \bibinfo{year}{2018}\natexlab{b}.
\newblock \showarticletitle{Rethinking Spatiotemporal Feature Learning:
  Speed-Accuracy Trade-offs in Video Classification}. In
  \bibinfo{booktitle}{\emph{European Conference of Computer Vision, {ECCV}}}
  \emph{(\bibinfo{series}{Lecture Notes in Computer Science},
  Vol.~\bibinfo{volume}{11219})}. \bibinfo{publisher}{Springer},
  \bibinfo{pages}{318--335}.
\newblock


\bibitem[\protect\citeauthoryear{Yu, Twardowski, Liu, Herranz, Wang, Cheng,
  Jui, and Weijer}{Yu et~al\mbox{.}}{2020}]%
        {yu2020semantic}
\bibfield{author}{\bibinfo{person}{Lu Yu}, \bibinfo{person}{Bartlomiej
  Twardowski}, \bibinfo{person}{Xialei Liu}, \bibinfo{person}{Luis Herranz},
  \bibinfo{person}{Kai Wang}, \bibinfo{person}{Yongmei Cheng},
  \bibinfo{person}{Shangling Jui}, {and} \bibinfo{person}{Joost van~de
  Weijer}.} \bibinfo{year}{2020}\natexlab{}.
\newblock \showarticletitle{Semantic drift compensation for class-incremental
  learning}. In \bibinfo{booktitle}{\emph{Proceedings of the IEEE/CVF
  Conference on Computer Vision and Pattern Recognition}}.
  \bibinfo{pages}{6982--6991}.
\newblock


\bibitem[\protect\citeauthoryear{Zhao, Fu, Kang, Tian, Wu, and Li}{Zhao
  et~al\mbox{.}}{2020}]%
        {zhao2020mgsvf}
\bibfield{author}{\bibinfo{person}{Hanbin Zhao}, \bibinfo{person}{Yongjian Fu},
  \bibinfo{person}{Mintong Kang}, \bibinfo{person}{Qi Tian},
  \bibinfo{person}{Fei Wu}, {and} \bibinfo{person}{Xi Li}.}
  \bibinfo{year}{2020}\natexlab{}.
\newblock \showarticletitle{MgSvF: Multi-Grained Slow vs. Fast Framework for
  Few-Shot Class-Incremental Learning}.
\newblock \bibinfo{journal}{\emph{arXiv preprint arXiv:2006.15524}}
  (\bibinfo{year}{2020}).
\newblock


\bibitem[\protect\citeauthoryear{Zhao, Wang, Fu, Wu, and Li}{Zhao
  et~al\mbox{.}}{2021}]%
        {zhao2021memory}
\bibfield{author}{\bibinfo{person}{Hanbin Zhao}, \bibinfo{person}{Hui Wang},
  \bibinfo{person}{Yongjian Fu}, \bibinfo{person}{Fei Wu}, {and}
  \bibinfo{person}{Xi Li}.} \bibinfo{year}{2021}\natexlab{}.
\newblock \showarticletitle{Memory Efficient Class-Incremental Learning for
  Image Classification}.
\newblock \bibinfo{journal}{\emph{IEEE Transactions on Neural Networks and
  Learning Systems}} (\bibinfo{year}{2021}).
\newblock


\bibitem[\protect\citeauthoryear{Zhao, Xiong, and Lin}{Zhao
  et~al\mbox{.}}{2018}]%
        {zhao2018trajectory}
\bibfield{author}{\bibinfo{person}{Yue Zhao}, \bibinfo{person}{Yuanjun Xiong},
  {and} \bibinfo{person}{Dahua Lin}.} \bibinfo{year}{2018}\natexlab{}.
\newblock \showarticletitle{Trajectory convolution for action recognition}. In
  \bibinfo{booktitle}{\emph{Proceedings of the 32nd International Conference on
  Neural Information Processing Systems}}. \bibinfo{pages}{2208--2219}.
\newblock


\bibitem[\protect\citeauthoryear{Zhou, Andonian, Oliva, and Torralba}{Zhou
  et~al\mbox{.}}{2018a}]%
        {zhou2018temporal}
\bibfield{author}{\bibinfo{person}{Bolei Zhou}, \bibinfo{person}{Alex
  Andonian}, \bibinfo{person}{Aude Oliva}, {and} \bibinfo{person}{Antonio
  Torralba}.} \bibinfo{year}{2018}\natexlab{a}.
\newblock \showarticletitle{Temporal relational reasoning in videos}. In
  \bibinfo{booktitle}{\emph{Proceedings of the European Conference on Computer
  Vision (ECCV)}}. \bibinfo{pages}{803--818}.
\newblock


\bibitem[\protect\citeauthoryear{Zhou, Sun, Zha, and Zeng}{Zhou
  et~al\mbox{.}}{2018b}]%
        {DBLP:conf/cvpr/ZhouSZZ18}
\bibfield{author}{\bibinfo{person}{Yizhou Zhou}, \bibinfo{person}{Xiaoyan Sun},
  \bibinfo{person}{Zheng{-}Jun Zha}, {and} \bibinfo{person}{Wenjun Zeng}.}
  \bibinfo{year}{2018}\natexlab{b}.
\newblock \showarticletitle{MiCT: Mixed 3D/2D Convolutional Tube for Human
  Action Recognition}. In \bibinfo{booktitle}{\emph{{IEEE} Conference on
  Computer Vision and Pattern Recognition, {CVPR}}}. \bibinfo{publisher}{{IEEE}
  Computer Society}, \bibinfo{pages}{449--458}.
\newblock


\bibitem[\protect\citeauthoryear{Zolfaghari, Singh, and Brox}{Zolfaghari
  et~al\mbox{.}}{2018}]%
        {DBLP:conf/eccv/ZolfaghariSB18}
\bibfield{author}{\bibinfo{person}{Mohammadreza Zolfaghari},
  \bibinfo{person}{Kamaljeet Singh}, {and} \bibinfo{person}{Thomas Brox}.}
  \bibinfo{year}{2018}\natexlab{}.
\newblock \showarticletitle{{ECO:} Efficient Convolutional Network for Online
  Video Understanding}. In \bibinfo{booktitle}{\emph{European Conference of
  Computer Vision, {ECCV}}} \emph{(\bibinfo{series}{Lecture Notes in Computer
  Science}, Vol.~\bibinfo{volume}{11206})}. \bibinfo{publisher}{Springer},
  \bibinfo{pages}{713--730}.
\newblock


\end{thebibliography}

\end{document}